\documentclass[10pt,twocolumn,letterpaper]{article}

\usepackage{cvpr}              

\usepackage{graphicx}
\usepackage{amsmath}
\usepackage{amssymb}
\usepackage{booktabs}
\usepackage{enumitem}
\usepackage{multirow}
\usepackage{wrapfig}
\usepackage{makecell}
\usepackage[accsupp]{axessibility}
\usepackage{footnote}

\usepackage{algorithm}
\usepackage[noend]{algcompatible}
\usepackage{arydshln} 
\usepackage[
  separate-uncertainty = true,
  multi-part-units = repeat
]{siunitx}

\usepackage[pagebackref,breaklinks,colorlinks]{hyperref}

\usepackage[capitalize]{cleveref}
\crefname{section}{Sec.}{Secs.}
\Crefname{section}{Section}{Sections}
\Crefname{table}{Table}{Tables}
\crefname{table}{Tab.}{Tabs.}

\newcommand{\nickname}{GrowSP}

\def\cvprPaperID{9090} 
\def\confName{CVPR}
\def\confYear{2023}

\begin{document}



\title{\nickname{}: Unsupervised Semantic Segmentation of 3D Point Clouds}

\author{Zihui Zhang, \quad Bo Yang\footnotemark[1], \quad Bing Wang, \quad Bo Li \\
Shenzhen Research Institute, The Hong Kong Polytechnic University\\ 
vLAR Group, The Hong Kong Polytechnic University\\
{\tt\small zihui.zhang@connect.polyu.hk, bo.yang@polyu.edu.hk}}

\twocolumn[{%
\renewcommand\twocolumn[1][]{#1}%
    \maketitle
    \begin{center}
        \vspace{-25pt}
        \centering
        \includegraphics[scale=0.35]{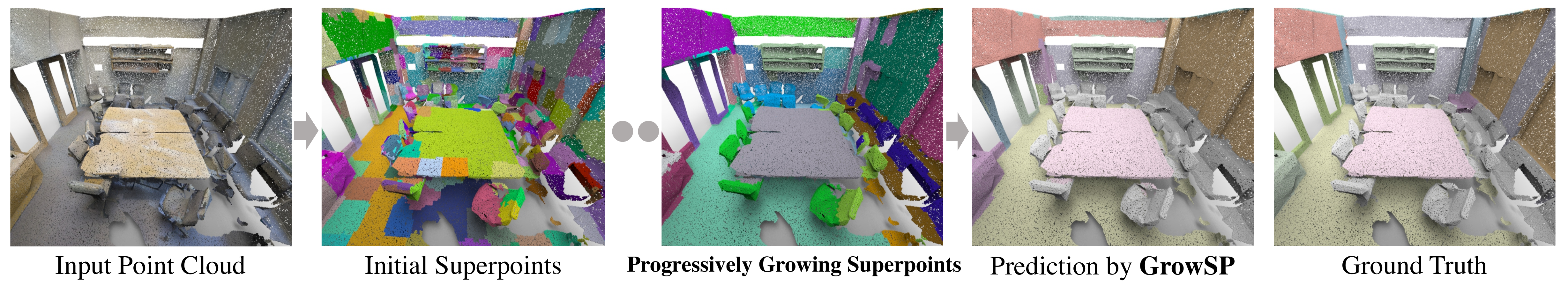}
        \vspace{-20pt}
        \captionof{figure}{Given an input point cloud with complex structures from S3DIS dataset \cite{Armeni2017}, our \nickname{} automatically discovers accurate semantic classes simply by progressively growing superpoints, without needing any human labels in training.}
        \label{fig:opening}
        \vspace{0pt}
    \end{center}
}]

\renewcommand{\thefootnote}{\fnsymbol{footnote}}
\footnotetext[1]{ Corresponding author}

\begin{abstract}
We study the problem of 3D semantic segmentation from raw point clouds. Unlike existing methods which primarily rely on a large amount of human annotations for training neural networks, we propose the first purely unsupervised method, called \textbf{\nickname{}}, to successfully identify complex semantic classes for every point in 3D scenes, without needing any type of human labels or pretrained models. The key to our approach is to discover 3D semantic elements via progressive growing of superpoints. Our method consists of three major components, 1) the feature extractor to learn per-point features from input point clouds, 2) the superpoint constructor to progressively grow the sizes of superpoints, and 3) the semantic primitive clustering module to group superpoints into semantic elements for the final semantic segmentation. We extensively evaluate our method on multiple datasets, demonstrating superior performance over all unsupervised baselines and approaching the classic fully-supervised PointNet. We hope our work could inspire more advanced methods for unsupervised 3D semantic learning.  
\end{abstract}\vspace{-0.5cm}

\section{Introduction}
Giving machines the ability to automatically discover semantic compositions of complex 3D scenes is crucial for many cutting-edge applications. 
In the past few years, there has been tremendous progress in fully-supervised semantic segmentation for 3D point clouds \cite{Guo2020}. From the seminar works PointNet \cite{Qi2016} and SparseConv \cite{Graham2018} to a plethora of recent neural models \cite{Qi2017,Li2018f,Wang2018c,Thomas2019,Hu2020}, both the accuracy and efficiency of per-point semantic estimation have been greatly improved. Unarguably, the success of these methods primarily relies on large-scale human annotations for training deep neural networks. However, manually annotating real-world 3D point clouds is extremely costly due to the unstructured data format \cite{Behley2019,Hu2020b}. To alleviate this problem, a number of recent methods start to use fewer 3D point labels \cite{Xu2020a,Hu2021}, cheaper 2D image labels \cite{Wang2019g,Zhu2021}, or active annotations \cite{Wu2021,Hu2022} in training. Although achieving promising results, they still need tedious human efforts to annotate or align 3D points across images for particular datasets, thus being inapplicable to novel scenes without training labels. 
\vspace{-0.25cm}

In this paper, we make the first step towards unsupervised 3D semantic segmentation of real-world point clouds. To tackle this problem, there could be two strategies: 1) to na\"ively adapt existing unsupervised 2D semantic segmentation techniques \cite{Caron2018,Ji2019,Cho2021} to 3D domain, and 2) to apply existing self-supervised 3D pretraining techniques \cite{PC,CSC} to learn discriminative per-point features followed by classic clustering methods to obtain semantic categories. 
For unsupervised 2D semantic methods, although achieving encouraging results on color images, they can be hardly extended to 3D point clouds primarily because: a) there is no general pretrained backbone to extract high-quality features for point clouds due to the lack of representative 3D datasets akin to ImageNet \cite{Russakovsky2015} or COCO \cite{Lin2014}, b) they are usually designed to group pixels with similar low-level features, \eg{} colors or edges, as a semantic class, whereas such a heuristic is normally not satisfied in 3D point clouds due to point sparsity and spatial occlusions. For self-supervised 3D pretraining methods, although the pretrained per-point features could be discriminative, they are lack of semantic meanings fundamentally because the commonly adopted data augmentation techniques do not explicitly capture categorical information. Section \ref{sec:exp} clearly demonstrates that all these methods fail catastrophically on 3D point clouds. 

Given a sparse point cloud composed of multiple semantic categories, we can easily observe that a relative small local point set barely contains distinctive semantic information. Nevertheless, when the size of a local point set is gradually growing, that surface patch naturally emerges as a basic element or primitive for a particular semantic class, and then it becomes much easier for us to identify the categories just by combining those basic primitives. For example, two individual 3D points sampled from a spacious room are virtually meaningless, whereas two patches might be easily identified as the \textit{back} and/or \textit{arm} of chairs. 

Inspired by this, we introduce a simple yet effective pipeline to automatically discover per-point semantics, simply by progressively growing the size of per-point neighborhood, without needing any human labels or pretrained backbone. In particular, our architecture consists of three major components: 1) a per-point feature extractor which is flexible to adopt an existing (untrained) neural network such as the powerful SparseConv \cite{Graham2018}; 2) a superpoint constructor which progressively creates larger and larger superpoints during training to guide semantic learning; 3) a semantic primitive clustering module which aims to group basic elements of semantic classes via an existing clustering algorithm such as K-means. The key to our pipeline is the superpoint constructor together with a progressive growing strategy in training. Basically, this component drives the feature extractor to progressively learn similar features for 3D points within a particular yet \textbf{grow}ing \textbf{s}uper\textbf{p}oint, while the features of different superpoints tend to be pushed as distinct elements of semantic classes. Our method is called \textbf{\nickname{}} and Figure \ref{fig:opening} shows qualitative results of an indoor 3D scene. Our contributions are:
\begin{itemize}[leftmargin=*]
\setlength{\itemsep}{1pt}
\setlength{\parsep}{1pt}
\setlength{\parskip}{1pt}
    \item We introduce the first purely unsupervised 3D semantic segmentation pipeline for real-world point clouds, without needing any pretrained models or human labels. 
    \item We propose a simple strategy to progressively grow superpoints during network training, allowing meaningful semantic elements to be learned gradually. 
    \item We demonstrate promising semantic segmentation results on multiple large-scale datasets, being clearly better than baselines adapted from unsupervised 2D methods and self-supervised 3D pretraining methods. Our code is at: \url{https://github.com/vLAR-group/GrowSP}
\end{itemize}

\section{Related Works}
\textbf{Learning with Strong Supervision:} With the advancement of 3D scanners, acquiring point clouds becomes easier and cheaper. In past five years, the availability of large-scale human-annotated point cloud datasets \cite{Armeni2017,Dai2017,Hackel2017,Behley2019,Varney2020,Tan2020,Hu2020b,Lin2022} enables fully-supervised neural methods to achieve remarkable 3D semantic segmentation results. These methods generally include: 1) 2D projection-based methods \cite{Wu2017e,Milioto2019,Kundu2020,Cortinhal2020} which project raw point clouds onto 2D images followed by mature 2D neural architectures to learn semantics; 2) voxel-based methods \cite{Graham2018,Meng2019,Choy2019,Lei2019,Zhu2021b} which usually voxelize unstructured point clouds into regular spheres, cubes, or cylinders followed by existing convolutional networks; 3) point-based methods \cite{Thomas2019,Wu2019,Liu2019h,Qi2017,Li2018f,Hu2020,Zhao2021a,Guo2021} which primarily follows the seminal PointNet \cite{Qi2016} to directly learn per-point features using shared MLPs. The performance of these methods can be further improved by the successful self-supervised pre-training techniques in recent studies \cite{Rao2020,Chen2021,Wang2021,Thabet2019,Xie2020,Zhang2021,Huang2021,Hou2021,Yin2022,Pang2022}. Although achieving excellent accuracy on existing benchmarks, they require densely-annotated 3D data for training. This is extremely costly and prohibitive in real applications. 

\textbf{Learning with Weak Supervision:} To alleviate the cost of human annotations, a number of works have started to learn 3D semantics using fewer or cheaper human labels in training. These weak labels primarily include: 1) fewer 3D point labels \cite{Xu2020a,Zhang2021a,Liu2021,Hu2021,Liu2022,Unal2022,Shi2022,Wu2022}, and 2) sub-cloud/seg-level/scene-level labels \cite{Wei2020,Tao2020,Ren2021,Chibane2022,Liu2022a}. The performance of these weakly-supervised methods can also be boosted by self-supervised pre-training techniques \cite{Hou2021,Xie2020,Zhang2021,Zhang2022}. Apart from these weak labels, supervision signals can also come from other domains such as labeled 2D images \cite{Liu2020c,Zhuang2021,Yan2022,Xu2022,Shin2022,Robert2022} or pretrained language models \cite{Zhang2022a,Ha2022,Rozenberszki2022}. Although obtaining encouraging results, these methods still need tedious human efforts to annotate or align data points. 
Fundamentally, they still cannot automatically discover semantic classes.

\begin{figure*}[t]
\setlength{\abovecaptionskip}{ 1 pt}
\setlength{\belowcaptionskip}{ -12 pt}
\centering
   \includegraphics[width=.85\linewidth]{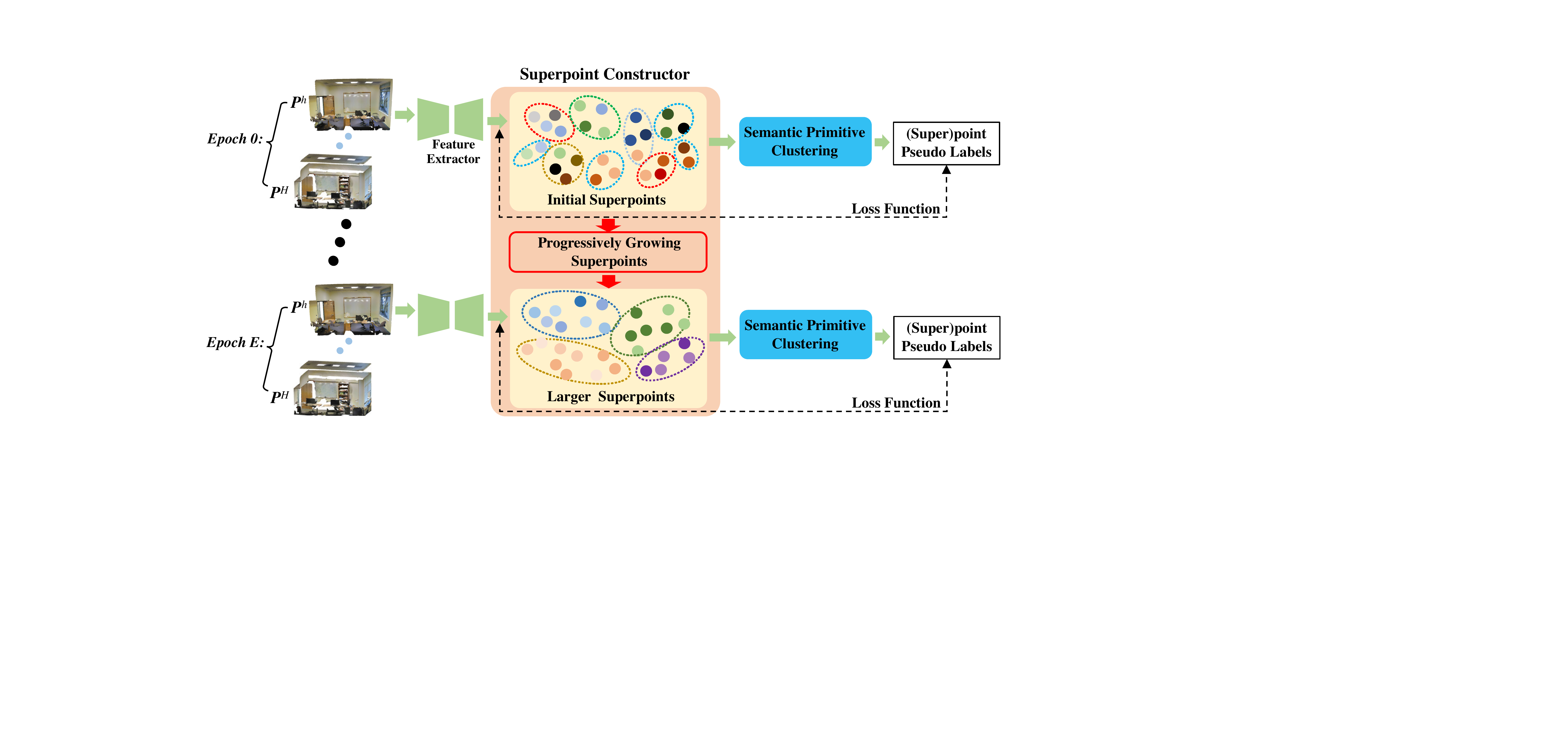}
\caption{The general learning framework of our \nickname{}. It mainly consists of three components: 1) the feature extractor which learns per-point features, 2) the superpoint constructor which progressively grows the sizes of superpoints, and 3) the semantic primitive clustering module which aims to group superpoints into semantic elements. \textbf{Note that, the superpoint constructor is no longer needed in testing}.}
\label{fig:meth_arch}
\end{figure*}

\textbf{Unsupervised Semantic Learning:} The work \cite{Sauder2019} learns point semantics by recovering voxel positions after randomly shuffling 3D points, and Canonical Capsules \cite{Sun2020} learns to decompose point clouds into object parts via self-canonicalization. However, both of them can only work on simple object point clouds. Technically, existing self-supervised 3D pretraining techniques \cite{PC,CSC} can be used for unsupervised semantic segmentation, just by learning discriminative per-point features followed by clustering. However, as shown in Section \ref{sec:exp_ssl}, the pretrained point features are actually lack of semantic meanings and fail to be grouped as classes. For 2D images, a number of recent works \cite{Caron2018,Ji2019,Ouali2020,Gansbeke2021,Cho2021} tackle the problem of unsupervised 2D semantic segmentation. However, due to the domain gap between images and point clouds, there is no existing work showing their applicability in 3D space. In fact, as demonstrated in Section \ref{sec:exp}, both the representative 2D methods IIC \cite{Ji2019} and PICIE \cite{Cho2021} fail catastrophically on point clouds, while our method achieves significantly better accuracy. 


\section{\nickname{}}
\subsection{Overview}

Our method generally formulates the problem of unsupervised 3D semantic segmentation as joint 3D point feature learning and clustering in the absence of human labels. As shown in Figure \ref{fig:meth_arch}, from a dataset with $H$ point clouds $\{\boldsymbol{P}^1 \cdots \boldsymbol{P}^h \cdots \boldsymbol{P}^H\}$, given one single scan $\boldsymbol{P}^h$ with $N$ points as input, \ie{}, $\boldsymbol{P}^h\in \mathbb{R}^{N\times 6}$ where each point has a location $\{x,y,z\}$ with color if available, the \textbf{feature extractor} firstly obtains per-point features $\boldsymbol{F}^h\in \mathbb{R}^{N\times K}$ where the embedding length $K$ is free to predefine, \eg{} $K=128$. We simply adopt the powerful SparseConv architecture \cite{Graham2018} without any pretraining step as our feature extractor. Implementation details are in Appendix.

Having the input point cloud $\boldsymbol{P}^h$ and its point features $\boldsymbol{F}^h$ at hand which are not meaningful in the very beginning, we will then feed them into our \textbf{superpoint constructor} to progressively generate larger and larger superpoints over more and more training epochs, as detailed in Section \ref{sec:meth_constructor}. These superpoints will be fed into our \textbf{semantic primitive clustering module}, generating pseudo labels for all superpoints, as discussed in Section \ref{sec:meth_cluster}. During training, these pseudo labels will be used to optimize the feature extractor.

\subsection{Superpoint Constructor}\label{sec:meth_constructor}

This module is designed to divide each input point cloud into pieces, such that each piece as whole ideally belongs to the same category. Intuitively, compared with individual points, a single piece is more likely to have geometric meanings, thus being easier to extract high-level semantics. In order to construct high-quality superpoints and aid the network to automatically discover semantics, here we ask two key questions: 
\begin{itemize}[leftmargin=*]
\setlength{\itemsep}{1pt}
\setlength{\parsep}{1pt}
\setlength{\parskip}{1pt}
\item First, what strategy should we use to construct superpoints? Naturally, if a superpoint keeps small, it can be highly homogeneous but lack of semantics. On the other hand, a larger superpoint may have better semantics but is error-prone if not constructed properly. In this regard, we propose to gradually grow the size of superpoints from small to large over more and more training epochs. 
\item Second, how to partition a point cloud into satisfactory pieces at the beginning, such that the network training can be bootstrapped effectively? Considering that point neural features are virtually meaningless in the early stage of network training, it is more reliable to simply leverage classic algorithms to obtain initial superpoints based on geometric features, \eg{} surface normal or connectivity. 
\end{itemize}
With these insights, we introduce the following mechanism to construct superpoints.

\textbf{Initial Superpoints:} As shown in the yellow block of Figure \ref{fig:meth_arch}, at the beginning of network training, the initial superpoints are constructed by VCCS \cite{Papon2013} followed by a region growing algorithm \cite{Adams1994}. They jointly take into account the spatial/normal/normalized RGB distances between 3D points. For a specific input point cloud $\boldsymbol{P}^h$, its initial superpoints are denoted as $\{\boldsymbol{\Tilde{p}}^h_1 \cdots \boldsymbol{\Tilde{p}}^h_{m^0} \cdots \boldsymbol{\Tilde{p}}^h_{M^0}\}$ where each superpoint $\boldsymbol{\Tilde{p}}^h_{m^0}$ consists of a small subset of original point cloud $\boldsymbol{P}^h$. Note that, for different point clouds, the number of their initial superpoints $M^0$ are usually different. Implementation details are in Appendix. Figure \ref{fig:meth_bootSP} shows an example of initial superpoints for an indoor room. 
\begin{figure}[t]
\setlength{\abovecaptionskip}{ 2 pt}
\setlength{\belowcaptionskip}{ -10 pt}
\centering
   \includegraphics[width=0.85\linewidth]{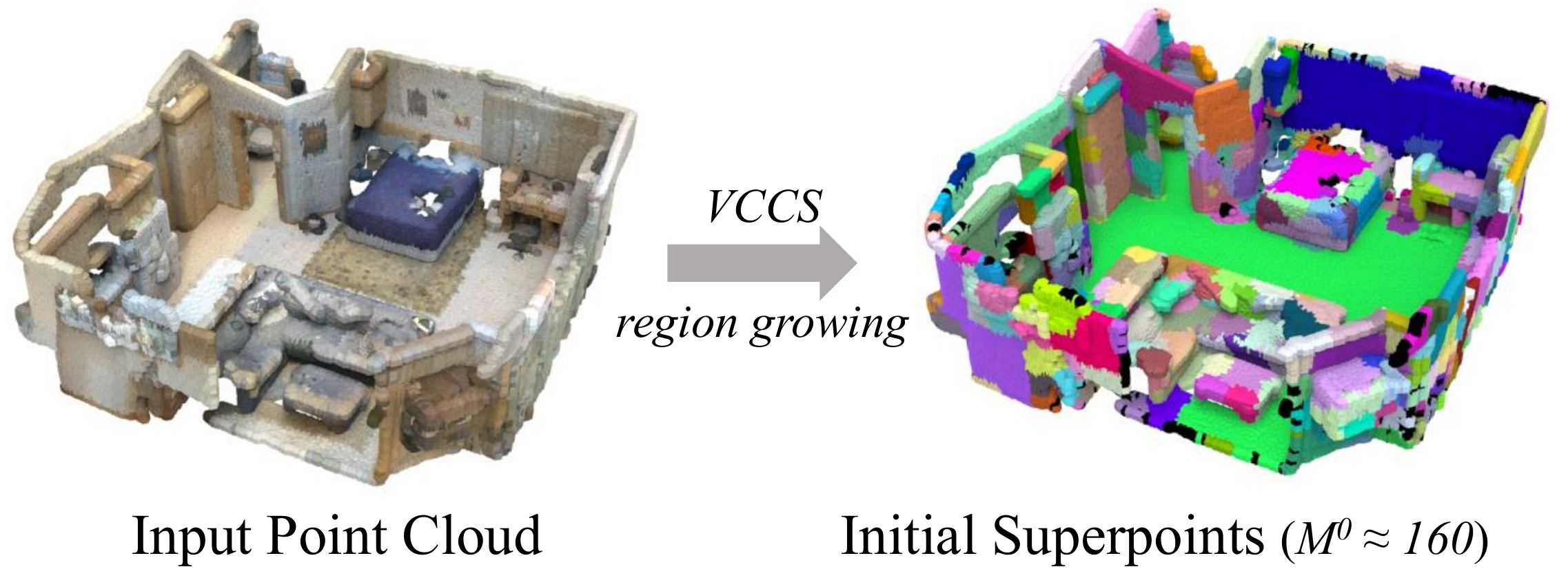}
\caption{An example of initial superpoints constructed by VCCS and region growing. Each colored patch represents a superpoint.}
\label{fig:meth_bootSP}
\end{figure}

\begin{figure}[ht]
\vspace{-0.3cm}
\setlength{\abovecaptionskip}{ 2 pt}
\setlength{\belowcaptionskip}{ -10 pt}
\centering
   \includegraphics[width=0.85\linewidth]{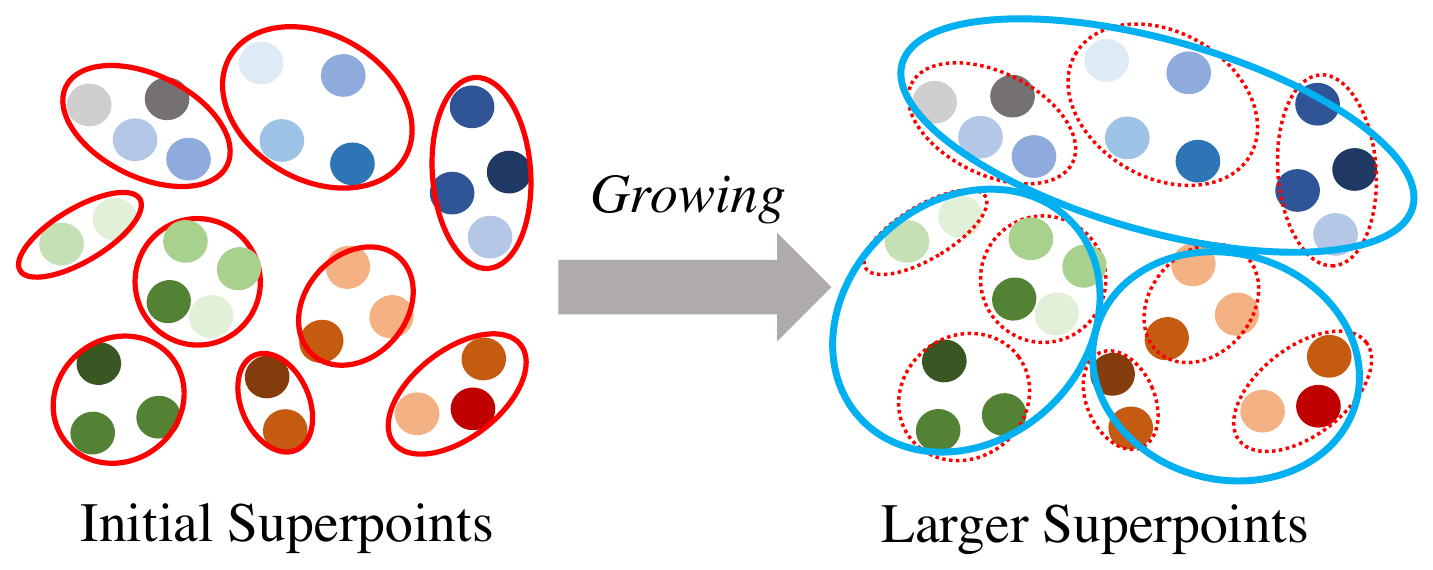}
\caption{An illustration of progressively growing superpoints.}
\label{fig:meth_growSP}
\end{figure}

\textbf{Progressively Growing Superpoints during Training:} Assuming the feature extractor is trained for epochs  using Algorithm \ref{alg:training} which will be detailed in Section \ref{sec:meth_imp}, the per-point features are expected to be more meaningful. In this regard, we turn to primarily use neural features to progressively construct larger superpoints for future training. As illustrated in Figure \ref{fig:meth_growSP}, each dot represents the neural embedding of a 3D point, and a red circle indicates an initial superpoint. The blue circle represents a larger superpoint absorbing one or multiple initial superpoints.  

In particular, for a specific input point cloud $\boldsymbol{P}^h$, we have its neural features $\boldsymbol{F}^h\in \mathbb{R}^{N\times K}$ and initial superpoints $\{\boldsymbol{\Tilde{p}}^h_1 \cdots \boldsymbol{\Tilde{p}}^h_{m^0} \cdots \boldsymbol{\Tilde{p}}^h_{M^0}\}$. Firstly, we will compute the mean neural features for initial superpoints, denoted as $\{ \boldsymbol{\Tilde{f}}^h_1 \cdots \boldsymbol{\Tilde{f}}^h_{m^0} \cdots \boldsymbol{\Tilde{f}}^h_{M^0} \}$:
\begin{equation}\label{eq:meth_featSP}
    \boldsymbol{\Tilde{f}}^h_{m^0} = \frac{1}{Q}\sum_{q=1}^Q \boldsymbol{f}^h_q, \quad \boldsymbol{f}^h_q\in \mathbb{R}^{1\times K}
\end{equation}
where $Q$ is the total number of 3D points within an initial superpoint $\boldsymbol{\Tilde{p}}^h_{m^0}$, and $\boldsymbol{f}^h_q$ is the feature vector retrieved from $\boldsymbol{F}^h$ for the $q^{th}$ 3D point of the superpoint. 

Secondly, having these initial superpoint features, we simply use K-means to group the $M^0$ vectors into $M^1$ clusters, where $M^1 < M^0$. Each cluster represents a new and larger superpoint. In total, we get new superpoints:
\begin{equation*}
     \{\boldsymbol{\Tilde{p}}^h_{1} \cdots \boldsymbol{\Tilde{p}}^h_{m^1} \cdots \boldsymbol{\Tilde{p}}^h_{M^1}\} \xleftarrow{\text{Kmeans}} \{ \boldsymbol{\Tilde{f}}^h_1 \cdots \boldsymbol{\Tilde{f}}^h_{m^0} \cdots \boldsymbol{\Tilde{f}}^h_{M^0} \}
\end{equation*}
Note that this superpoint growing step is conducted independently on each input point cloud. The much smaller $M^1$, the more aggressive this growing step. 

After every a certain number of training epochs, \ie{} one round, we will compute the next level of larger superpoints by repeating above two steps. Given $T$ levels of growing, the number of superpoints for an input point cloud will be reduced from $M^1 \rightarrow M^2 \rightarrow M^t$ until to a small value $M^T$. In each epoch, all superpoints of the entire dataset will be fed into the semantic primitive clustering module.

\subsection{Semantic Primitive Clustering}\label{sec:meth_cluster}
For every epoch, each input point cloud will have a number of superpoints, each of which representing a particular part of objects or stuff. As to the whole dataset, all superpoints together can be regarded as a huge set of basic semantic elements or primitives, such as chair backs, table surfaces, \etc{}. In order to discover semantics from these superpoints, two issues need to be addressed:
\begin{itemize}[leftmargin=*]
\setlength{\itemsep}{1pt}
\setlength{\parsep}{1pt}
\setlength{\parskip}{1pt}
\item First, how to effectively group these superpoints? A straightforward way is to directly cluster all superpoints into a number of object categories using an existing clustering algorithm. However, we empirically find that this is excessively aggressive, because many superpoints belonging to different categories are similar and then wrongly assigned to the same semantic group at the early training stage, and it is hard to be corrected over time. In this regard, we opt to constantly group all superpoints into a relatively large number of clusters in all training epochs.
\item Second, are the neural features of superpoints discriminative enough for semantic clustering? Again, considering that the neural features of 3D points as well as superpoints are meaningless at the beginning of network training, it is more reliable to explicitly take into account point geometry features such as surface normal distributions to augment discrimination of superpoints. To this end, for each superpoint, we simply stack both its neural features and the classic PFH feature \cite{Rusu2008} for clustering. 
\end{itemize}

\begin{algorithm*}[th]
\caption{The training pseudocode of our \nickname{}. Given a dataset with $H$ point cloud scans $\{\boldsymbol{P}^1 \cdots \boldsymbol{P}^h \cdots \boldsymbol{P}^H \}$. $\hat{E}$ is a predefined number of epochs for periodically and progressively growing superpoints. 
The hyperparameters $M^1$ is set as 80, $S$ as 300, $\hat{E}$ as 10 in all experiments.}
\label{alg:training}
\begin{algorithmic} 
\STATE{\textit{Epoch 0: Initial superpoints construction.}} 
\STATE{\phantom{xx}$\bullet$} Apply VCCS \cite{Papon2013} followed by region growing algorithm \cite{Adams1994} on each point cloud scan, and obtain all initial superpoints:\\ 
\phantom{xx} $ \mathbb{\Tilde{P}}_0 =
\big( \{\boldsymbol{\Tilde{p}}^1_1 \cdots \boldsymbol{\Tilde{p}}^1_{m^0} \cdots \boldsymbol{\Tilde{p}}^1_{M^0} \} 
 \cdots \{\boldsymbol{\Tilde{p}}^h_1 \cdots \boldsymbol{\Tilde{p}}^h_{m^0} \cdots \boldsymbol{\Tilde{p}}^h_{M^0} \}
 \cdots \{\boldsymbol{\Tilde{p}}^H_1 \cdots \boldsymbol{\Tilde{p}}^H_{m^0} \cdots \boldsymbol{\Tilde{p}}^H_{M^0} \}
\big) 
$;
\STATE{\textit{\phantom{xxx}Note: the number of initial superpoints for different point cloud scans are usually different. We use the same $M^0$ to avoid an abuse of notation.}}\vspace{0.2cm}

\STATE{\phantom{xx}$\bullet$} Load superpoints $\mathbb{\Tilde{P}} \leftarrow \mathbb{\Tilde{P}}_0$ for training;
\STATE{\phantom{xx}$\bullet$} Predefine the number of superpoints at total $T$ levels: $\{M^1 \cdots M^t \cdots M^T\}$ where $M^0 > M^1 > M^t > M^T$; 
\STATE{\textit{\phantom{xxx}Note: for simplicity, we choose consecutive integers in all experiments, \ie{} $M^T = M^{T-1} -1 = \cdots = M^1 - (T-1)$.}}

\STATE{\phantom{xx}$\bullet$} Initilize superpoint level $t = 0$;
\FOR {training epoch $e$ in $\{1, 2, \cdots E\}$}{}
\STATE{\textbf{if $e \% \hat{E}$ != 0: To optimize the neural network}} 
\STATE{\phantom{xx}$\bullet$} Feed all $H$ point clouds into the feature extractor, and obtain neural features $\{\boldsymbol{F}^1 \cdots \boldsymbol{F}^h \cdots \boldsymbol{F}^H\}$;
\STATE{\phantom{xx}$\bullet$} Obtain neural features according to Equation \ref{eq:meth_featSP} and PFH features \cite{Rusu2008} for all superpoints $\mathbb{\Tilde{P}}$; 

\STATE{\phantom{xx}$\bullet$} Apply K-means to cluster all superpoints of the entire dataset into $S$ semantic primitives, where each superpoint and individual 3D points within it will be assigned a one-hot pseudo label;
\STATE{\phantom{xx}$\bullet$} The centroids of $S$ semantic primitives estimated by K-means are used as a classifier to classify all individual 3D points of the dataset. Cross-entropy loss is applied between the logits and pseudo labels to optimize the whole network. \vspace{0.2cm}

\STATE{\textbf{if $e \% \hat{E}$ == 0: To progressively grow superpoints}} 
\STATE{\phantom{xx}$\bullet$} Update superpoint level $t = t+1$, and get the corresponding predefined superpoint number $M^t$;
\STATE{\phantom{xx}$\bullet$} For each point cloud, obtain the latest neural features for each initial superpoint, and then apply K-means to cluster these initial superpoints into $M^t$ new superpoints. For simplicity, we use the same value $M^t$ for all $H$ point clouds: \\
\phantom{xxxx} $ \mathbb{\Tilde{P}}_t =
\big( \{\boldsymbol{\Tilde{p}}^1_1 \cdots \boldsymbol{\Tilde{p}}^1_{m^t} \cdots \boldsymbol{\Tilde{p}}^1_{M^t} \} 
 \cdots \{\boldsymbol{\Tilde{p}}^h_1 \cdots \boldsymbol{\Tilde{p}}^h_{m^t} \cdots \boldsymbol{\Tilde{p}}^h_{M^t} \}
 \cdots \{\boldsymbol{\Tilde{p}}^H_1 \cdots \boldsymbol{\Tilde{p}}^H_{m^t} \cdots \boldsymbol{\Tilde{p}}^H_{M^t} \}
\big) $;
\STATE{\phantom{xx}$\bullet$} Update superpoints $\mathbb{\Tilde{P}} \leftarrow \mathbb{\Tilde{P}}_t$
\ENDFOR
\end{algorithmic}
\end{algorithm*}

As shown in the blue block of Figure \ref{fig:meth_arch}, taking the first epoch as an example, given all $H$ point clouds in the whole dataset $\{\boldsymbol{P}^1 \cdots \boldsymbol{P}^H\}$, we have all initial superpoints
$
\big( \{\boldsymbol{\Tilde{p}}^1_1 \cdots \boldsymbol{\Tilde{p}}^1_{m^0} \cdot \cdot\} 
 \cdots \{\boldsymbol{\Tilde{p}}^H_1 \cdots \boldsymbol{\Tilde{p}}^H_{m^0} \cdot \cdot\}
\big)
$
and their features 
$\big( \{\boldsymbol{\hat{f}}^1_1 \cdots \boldsymbol{\hat{f}}^1_{m^0} \cdot \cdot\} 
 \cdots \{\boldsymbol{\hat{f}}^H_1 \cdots \boldsymbol{\hat{f}}^H_{m^0} \cdot \cdot\}
\big)
$. Each superpoint's features are geometry augmented:
\begin{equation}
    \boldsymbol{\hat{f}}^1_{m^0} = \boldsymbol{\Tilde{f}}^1_{m^0} \oplus \boldsymbol{\ddot{f}}^1_{m^0}
\end{equation}
where the neural features $\boldsymbol{\Tilde{f}}^1_{m^0}$ are obtained by Equation \ref{eq:meth_featSP} and concatenated with 10-dimensional PFH features $\boldsymbol{\ddot{f}}^1_{m^0}$. We simply adopt K-means to cluster all these superpoint features into $S$ semantic primitives:
\begin{equation*}
    \textit{$S$ primitives} \xleftarrow{\text{Kmeans}} \big( \{\boldsymbol{\hat{f}}^1_1 \cdots \boldsymbol{\hat{f}}^1_{m^0} \cdot \cdot\}
 \cdots \{\boldsymbol{\hat{f}}^H_1 \cdots \boldsymbol{\hat{f}}^H_{m^0} \cdot \cdot\}
\big)
\end{equation*}

\textbf{Loss Function:} Naturally, each superpoint and individual 3D points within it will be given an $S$-dimensional one-hot pseudo-label. For all $S$ primitives, we use the corresponding centroids (PFH simply dropped) estimated by K-means as a classifier to classify all individual 3D points, obtaining $S$-dimensional logits. Lastly, the standard cross-entropy loss is applied between logits and pseudo-labels to optimize the neural extractor from scratch. 

\subsection{Implementation}\label{sec:meth_imp}

\textbf{Training Phase:} To better illustrate our \nickname{}, Algorithm \ref{alg:training} clearly presents all steps of our pipeline during training. Notably, our method does not need to be given the actual number of semantic classes in training, because we simply learn semantic primitives. 

\textbf{Testing Phase:} Once the network is well-trained, we keep the centroids of $S$ semantic primitives estimated by K-means on training split. In testing, these centroids are directly grouped into $C$ semantic classes using K-means. The newly obtained centroids for the $C$ classes are used as the final classifier. Given a test point cloud, all per-point neural features are directly classified as one of $C$ classes, \textbf{without needing to construct superpoints anymore.} For the final evaluation metrics calculation, we follow \cite{Cho2021} to use Hungarian algorithm to match predicted classes with ground truth labels.
Implementation details are in Appendix.

\section{Experiments}\label{sec:exp}

Our method is mainly evaluated on two large-scale indoor datasets and one challenging outdoor LIDAR dataset: S3DIS \cite{Armeni2017}, ScanNet \cite{Dai2017} and SemanticKITTI \cite{Behley2019}. We also conduct cross-dataset experiments to evaluate the generalization ability on unseen scenes, and results are supplied in Appendix. For evaluation metrics, we report the standard mean Intersection-over-Union (mIoU), Overall Accuracy (OA), mean Accuracy (mAcc) of all classes. More quantitative and qualitative results are in Appendix.

\begin{table}[th] 
\centering
\setlength{\abovecaptionskip}{ 2 pt}
\setlength{\belowcaptionskip}{ -10 pt}
\caption{Quantitative results of our method and baselines on the Area-5 of S3DIS dataset \cite{Armeni2017}. Only 12 classes are evaluated. The performance standard deviations of unsupervised methods are calculated over the last five checkpoints during the final 50 epochs.}
\label{tab:exp_s3dis_area5}
\resizebox{0.48\textwidth}{!}
{
\begin{tabular}{crccc}
\toprule[1.0pt]
& & OA(\%) & mAcc(\%)& mIoU(\%) \\
\toprule[1.0pt]
\multirow{3}{*}{\makecell[c]{Supervised\\Methods} }& PointNet \cite{Qi2016}& 77.5 & 59.1 & 44.6 \\
& PointNet++ \cite{Qi2017}& 77.5 & 62.6 & 50.1 \\
& SparseConv \cite{Graham2018} & 88.4 & 69.2 & 60.8\\
\toprule[1.0pt]
\multirow{8}{*}{\makecell[c]{Unsupervised\\Methods} } 
& RandCNN    &23.3$\pm$2.6 &17.3$\pm$1.1 &9.2$\pm$1.2 \\

& van Kmeans &21.4$\pm$0.6  &21.2$\pm$1.6 &8.7$\pm$0.3 \\

& van Kmeans-S &21.9$\pm$0.5  &22.9$\pm$0.4 &9.0$\pm$0.2 \\

& van Kmeans-PFH &23.2$\pm$0.7  &23.6$\pm$1.7 &10.2$\pm$1.4\\

& van Kmeans-S-PFH &22.8$\pm$1.7  &20.6$\pm$0.7 &9.2$\pm$0.9 \\

& IIC \cite{Ji2019} &28.5$\pm$0.2  &12.5$\pm$0.2 &6.4$\pm$0 \\

& IIC-S \cite{Ji2019} &29.2$\pm$0.5  &13.0$\pm$0.2 &6.8$\pm$0 \\

& IIC-PFH \cite{Ji2019} &28.6$\pm$0.1  &16.8$\pm$0.1 &7.9$\pm$0.4 \\

& IIC-S-PFH \cite{Ji2019} &31.2$\pm$0.2  &16.3$\pm$0.1 &9.1$\pm$0.1 \\

& PICIE \cite{Cho2021} &61.6$\pm$1.5  &25.8$\pm$1.6 &17.9$\pm$0.9 \\

& PICIE-S \cite{Cho2021} &49.6$\pm$2.8  &28.9$\pm$1.0 &20.0$\pm$0.6 \\

& PICIE-PFH \cite{Cho2021} &54.0$\pm$0.8  &36.8$\pm$1.7 &24.4$\pm$0.6 \\

& PICIE-S-PFH \cite{Cho2021} &48.4$\pm$0.9  &40.4$\pm$1.6 &25.2$\pm$1.2 \\

& \textbf{\nickname{}(Ours)} &\textbf{78.4}$\pm$1.5 &\textbf{57.2}$\pm$1.7 &\textbf{44.5}$\pm$1.1 \\
\bottomrule[1.0pt]
\end{tabular}
}\vspace{-0.4cm}
\end{table}

\begin{table}[thb]
\centering
\setlength{\abovecaptionskip}{ 2 pt}
\setlength{\belowcaptionskip}{ -4 pt}
\caption{Quantitative results of 6-fold cross validation on S3DIS dataset \cite{Armeni2017}. Only 12 classes excluding \textit{clutter} are evaluated.}
\label{tab:exp_s3dis_6fold}
\resizebox{0.48\textwidth}{!}{
\begin{tabular}{crccc}
\toprule[1.0pt]
 & & OA(\%) & mAcc(\%) & mIoU(\%) \\
\toprule[1.0pt]
\multirow{3}{*}{\makecell[c]{Supervised\\Methods}} & PointNet \cite{Qi2016} & 75.9 & 67.1 & 49.4 \\
 & PointNet++ \cite{Qi2017} & 77.1 & 74.1 & 55.1 \\
 & SparseConv \cite{Graham2018} & 89.4 & 78.1 & 69.2 \\
\toprule[1.0pt]
\multirow{8}{*}{\makecell[c]{Unsupervised\\Methods}} & RandCNN & 23.1 & 18.4 & 9.3 \\
   & van Kmeans & 20.0 & 21.5 & 8.8 \\
   & van Kmeans-S & 20.0 & 22.3 & 8.8 \\
   & van Kmeans-PFH & 23.9 & 24.7 & 10.9 \\
   & van Kmeans-S-PFH & 23.4 & 20.8 & 9.5 \\
   & IIC \cite{Ji2019} & 32.8 & 14.7 & 8.5 \\
   & IIC-S \cite{Ji2019} & 29.4 & 15.1 & 7.7 \\
   & IIC-PFH \cite{Ji2019} & 29.5 & 13.2 & 6.7 \\
   & IIC-S-PFH \cite{Ji2019} & 26.3 & 13.6 & 7.2 \\
   & PICIE \cite{Cho2021} & 46.4 & 28.1 & 17.8 \\
   & PICIE-S \cite{Cho2021} & 50.7 & 30.8 & 21.6 \\
   & PICIE-PFH \cite{Cho2021} & 55.0 & 38.8 & 26.6 \\
   & PICIE-S-PFH \cite{Cho2021} & 49.1 & 40.5 & 26.7 \\
&\textbf{\nickname{} (Ours)} &\textbf{76.0} &\textbf{59.4} &\textbf{44.6} \\
\toprule[1.0pt]
\end{tabular}
}\vspace{-0.7cm}
\end{table}

\subsection{Evaluation on S3DIS} \label{sec:exp_s3dis}
The S3DIS dataset \cite{Armeni2017} consists of 6 large areas with 271 rooms. Each point belongs to one of 13 categories. We find that the \textit{clutter} class across different rooms does not have consistent geometric patterns and semantic meanings. In the absence of human labels, to automatically discover such diverse geometries as a common category is challenging and also unreasonable. Therefore, in the final testing stage, we only group all points except \textit{clutter} into 12 classes. The \textit{clutter} points are excluded (masked) for metrics calculation. Note that, in training, all points including \textit{clutter} are fed into the network, but \textit{clutter} points are not used for clustering and loss computation. We use the standard 6-fold cross validation in our experiments. $M^T$ is set as 20 which is slightly larger than the actual 12 semantic classes. 

Since there is no existing unsupervised method for semantic segmentation on 3D point clouds, we implement the following four baselines: 1) \textbf{RandomCNN} which uses K-means to directly cluster the per-point features into 12 classes, where the features are obtained from the randomly initialized backbone as ours; 2) \textbf{vanilla K-means} which directly clusters raw 3D points (\textit{xyzrgb}) into 12 classes; 3) \textbf{IIC} \cite{Ji2019} which is adapted from the existing unsupervised 2D method, where we change to use the same backbone as ours; 4) \textbf{PICIE} \cite{Cho2021} which is also adapted from 2D domain with the same backbone as ours. For an extensive comparison, we also adopt the same semantic primitive clustering module on baselines, denoted as van Kmeans-S/IIC-S/PICIE-S. The PFH features are also applied to them, denoted as van Kmeans-S-PFH/IIC-S-PFH/PICIE-S-PFH. Additionally, three classic fully-supervised methods PointNet \cite{Qi2016}, PointNet++ \cite{Qi2017}, and SparseConv \cite{Graham2018} are also included for comparison. These baselines are all carefully trained and evaluated using the same settings as ours. More details of implementation and experiments are in Appendix. 

\begin{figure*}[t]
\setlength{\abovecaptionskip}{ 0. pt}
\setlength{\belowcaptionskip}{ -15 pt}
\centering
   \includegraphics[width=.97\linewidth]{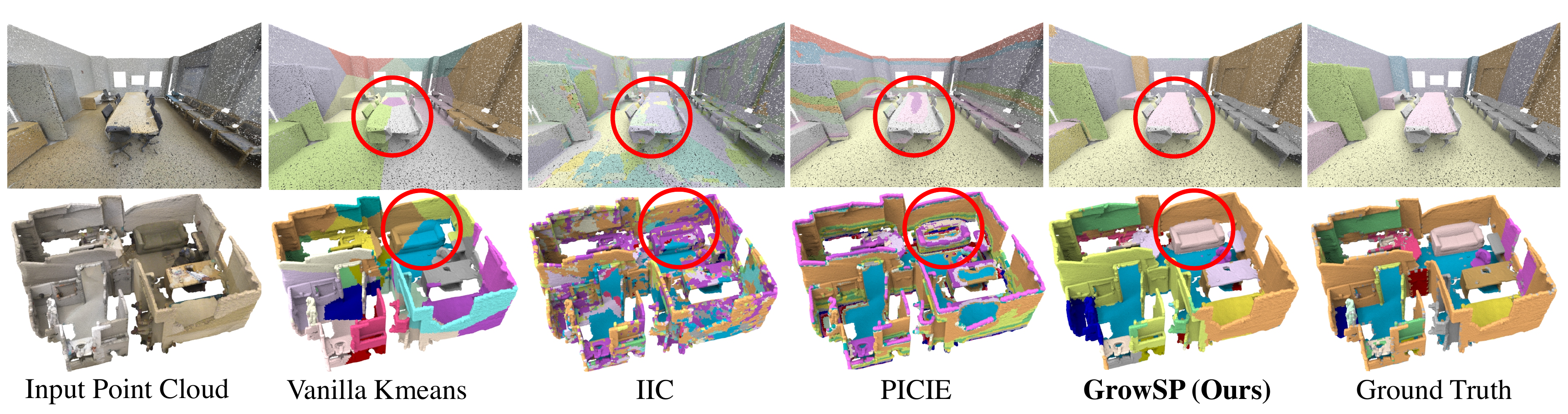}
\caption{Qualitative results of our method and baselines. The top row is from S3DIS dataset, the row below from ScanNet. Each color represent one semantic class. Red circles highlight the differences.}
\label{fig:exp_s3dis_scannet}
\end{figure*}

\textbf{Analysis:} As shown in Tables \ref{tab:exp_s3dis_area5}\&\ref{tab:exp_s3dis_6fold}, our method clearly outperforms all unsupervised baselines by large margins. Basically, the RandomCNN and vanilla K-means fail to obtain any meaningful semantic classes, due to the lack of meaningful point features. Despite using the same powerful SparseConv \cite{Graham2018} as feature extractor, neither IIC \cite{Ji2019} nor PICIE \cite{Cho2021} can obtain high-quality semantic segmentation results, primarily because neither takes full use of semantics emerging from larger and larger point regions, but instead tends to simply group similar points according to low-level features. Adding semantic primitive clustering module can  help PICIE get better results, but it is still significantly worse than our method. Figure \ref{fig:exp_s3dis_scannet} shows qualitative results.

\subsection{Evaluation on ScanNet}\label{sec:exp_scannet}

The ScanNet dataset \cite{Dai2017} has 1201 rooms for training, 312 rooms for offline validation, and 100 rooms for online hidden testing. Each point belongs to one of 20 object categories or undefined background. Since there is no unsupervised result on the hidden test set and uploading baseline results to online is not allowed, we turn to compare with the above 4 unsupervised baselines on the validation split. For this dataset, the hyperparameter $M^T$ is chosen as 30 which is slightly larger than 20 categories. During training, all raw point clouds are fed into neural networks, while the undefined points are not used for loss computation and clustering. They will also be masked out during testing. 


\begin{table}[thb]
\centering
\setlength{\belowcaptionskip}{ -6 pt}
\caption{Quantitative results on the \textbf{validation split} of ScanNet dataset \cite{Dai2017}. All 20 categories are evaluated.}
\vspace{-0.3cm}
\label{tab:exp_scannet_val}
\resizebox{0.48\textwidth}{!}{
\begin{tabular}{crccc}
\toprule[1.0pt]
 & & OA(\%) & mAcc(\%) & mIoU(\%) \\
\toprule[1.0pt]
\multirow{8}{*}{\makecell[c]{Unsupervised\\Methods}} & RandCNN & 11.9$\pm$0.4 & 8.4$\pm$0.1 & 3.2$\pm$0 \\
   & van Kmeans & 10.1$\pm$0.1 & 10.0$\pm$0.1 & 3.4$\pm$0 \\
   & van Kmeans-S & 10.2$\pm$0.1 & 9.8$\pm$0.3 & 3.4$\pm$0.1 \\
   & van Kmeans-PFH & 10.4$\pm$0.2 & 10.3$\pm$0.7 & 3.5$\pm$0.2 \\
   & van Kmeans-S-PFH & 12.2$\pm$0.6 & 9.3$\pm$0.5 & 3.6$\pm$0.1 \\
   & IIC \cite{Ji2019} & 27.7$\pm$2.7 & 6.1$\pm$1.2 & 2.9$\pm$0.8 \\
   & IIC-S \cite{Ji2019} & 18.3$\pm$2.6 & 6.7$\pm$0.6 & 3.4$\pm$0.1 \\
   & IIC-PFH \cite{Ji2019} & 25.4$\pm$0.1 & 6.3$\pm$0 & 3.4$\pm$0 \\
   & IIC-S-PFH \cite{Ji2019} & 18.9$\pm$0.3 & 6.3$\pm$0.2 & 3.0$\pm$0.1 \\
   & PICIE \cite{Cho2021} & 20.4$\pm$0.5 & 16.5$\pm$0.3 & 7.6$\pm$0 \\
   & PICIE-S \cite{Cho2021} & 35.6$\pm$1.1 & 13.7$\pm$1.5 & 8.1$\pm$0.5 \\
   & PICIE-PFH \cite{Cho2021} & 23.1$\pm$1.4 & 14.0$\pm$0.1 & 8.1$\pm$0.3 \\
   & PICIE-S-PFH \cite{Cho2021} & 23.6$\pm$0.4 & 15.1$\pm$0.6 & 7.4$\pm$0.2 \\
&\textbf{\nickname{} (Ours)} &\textbf{57.3}$\pm$2.3 & \textbf{44.2}$\pm$3.1 &\textbf{25.4}$\pm$2.3 \\
\toprule[1.0pt]
\end{tabular}
}\vspace{-0.4cm}
\end{table}

\begin{table}[thb]
\centering
\setlength{\abovecaptionskip}{ 2 pt}
\caption{Quantitative results on the \textbf{online hidden split} of ScanNet dataset \cite{Dai2017}. All 20 categories are evaluated.}
\label{tab:exp_scannet_online}
\resizebox{0.4\textwidth}{!}{
\begin{tabular}{crc}
\toprule[1.0pt]
 &  & mIoU(\%) \\
\toprule[1.0pt]
\multirow{3}{*}{\makecell[c]{Supervised Methods}} & PointNet++ \cite{Qi2017}  & 33.9 \\
 & DGCNN \cite{Wang2018c} & 44.6 \\
 & PointCNN \cite{Li2018f} & 45.8 \\
 & SparseConv \cite{Graham2018}  & 72.5 \\
\toprule[1.0pt] 
\multirow{1}{*}{\makecell[c]{Unsupervised Method}}
&\textbf{\nickname{} (Ours)}  & 26.9 \\ 
\toprule[1.0pt]
\end{tabular}
}\vspace{-0.4cm}
\end{table}

\textbf{Analysis:} As shown in Table \ref{tab:exp_scannet_val}, all unsupervised baselines fail on this challenging dataset. IIC and PICIE are just slightly better than RandCNN, demonstrating the level of segmentation difficulty. By contrast, our \nickname{} achieves very encouraging results. As shown in Table \ref{tab:exp_scannet_online}, our method obtains a similar mIoU score on the online benchmark.  
Qualitative results are presented in Figure \ref{fig:exp_s3dis_scannet}.

\subsection{Evaluation on SemanticKITTI}
\label{sec:exp_semantickitti}

The SemanticKITTI dataset \cite{Behley2019} consists of 21 sequences of 43552 outdoor LIDAR scans. It has 19130 scans for training, 4071 for validation and 20351 for online testing. Each point belongs to one of 19 semantic categories or undefined background. 
We set $M^T$ as 30, and exclude(masked) undefined background points during testing, and ignore them in training. Detailed implementations and results for all categories are in appendix.

\begin{table}[thb]
\centering
\setlength{\belowcaptionskip}{ -10 pt}
\caption{Quantitative results on the \textbf{validation split} of SemanticKITTI dataset \cite{Behley2019}. All 19 categories are evaluated.}
\vspace{-0.3cm}
\label{tab:exp_semantickitti_val}
\resizebox{0.48\textwidth}{!}{
\begin{tabular}{crccc}
\toprule[1.0pt]
 & & OA(\%) & mAcc(\%) & mIoU(\%) \\
\toprule[1.0pt]
\multirow{8}{*}{\makecell[c]{Unsupervised\\Methods}} 
   & RandCNN & 25.4$\pm$3.3 & 6.0$\pm$0.2 & 3.2$\pm$0.1 \\
   & van Kmeans & 8.1$\pm$0 & 8.2$\pm$0.1 & 2.4$\pm$0 \\
   & van Kmeans-S & 10.3$\pm$0.3 & 7.7$\pm$0.1 & 2.6$\pm$0 \\
   & van Kmeans-PFH & 11.2$\pm$0.6 & 7.5$\pm$0.7 & 2.7$\pm$0.1 \\
   & van Kmeans-S-PFH & 13.2$\pm$1.8 & 8.1$\pm$0.4 & 3.0$\pm$0.2 \\
   & IIC \cite{Ji2019} & 26.2$\pm$1.5 & 5.8$\pm$0.4 & 3.1$\pm$0.3 \\
   & IIC-S \cite{Ji2019} & 23.9$\pm$1.1 & 6.1$\pm$0.3 & 3.2$\pm$0.2 \\
   & IIC-PFH \cite{Ji2019} & 20.1$\pm$0.1 & 7.2$\pm$0.1 & 3.6$\pm$0 \\
   & IIC-S-PFH \cite{Ji2019} & 23.4$\pm$0 & 9.0$\pm$0 & 4.6$\pm$0 \\
   & PICIE \cite{Cho2021} & 22.3$\pm$0.4 & 14.6$\pm$0.3 & 5.9$\pm$0.1 \\
   & PICIE-S \cite{Cho2021} & 18.4$\pm$0.5 & 13.2$\pm$0.2 & 5.1$\pm$0.1 \\
   & PICIE-PFH \cite{Cho2021} & 46.6$\pm$0.2 & 10.1$\pm$0 & 4.7$\pm$0 \\
   & PICIE-S-PFH \cite{Cho2021} & 42.7$\pm$2.1 & 11.5$\pm$0.2 & 6.8$\pm$0.6 \\
&\textbf{\nickname{} (Ours)} &\textbf{38.3}$\pm$1.0 & \textbf{19.7}$\pm$0.6 &\textbf{13.2}$\pm$0.1 \\
\toprule[1.0pt]
\end{tabular}
}\vspace{-0.4cm}
\end{table}

\begin{table}[thb]
\centering
\setlength{\belowcaptionskip}{ -10 pt}
\caption{Quantitative results on the \textbf{online hidden split} of SemanticKITTI dataset \cite{Behley2019}. All 19 categories are evaluated.}
\vspace{-0.3cm}
\label{tab:exp_semantickitti_online}
\resizebox{0.4\textwidth}{!}{
\begin{tabular}{crc}
\toprule[1.0pt]
 &  & mIoU(\%) \\
\toprule[1.0pt]
\multirow{3}{*}{\makecell[c]{Supervised Methods}} & PointNet \cite{Qi2016}  & 14.6 \\
 & PointNet++ \cite{Qi2017} & 20.1 \\
 & SparseConv \cite{Graham2018}  & 53.2 \\
\toprule[1.0pt] 
\multirow{1}{*}{\makecell[c]{Unsupervised Method}}
&\textbf{\nickname{} (Ours)}  & 14.3 \\ 
\toprule[1.0pt]
\end{tabular}
}\vspace{-0.2cm}
\end{table}

\textbf{Analysis:} As shown in Tables \ref{tab:exp_semantickitti_val}\&\ref{tab:exp_semantickitti_online}, all unsupervised baselines fail on this outdoor dataset. Our method achieves satisfactory segmentation results on par with the fully-supervised PointNet\cite{Qi2016}. Nevertheless, our method cannot automatically discover the minor classes such as \textit{bike, cyclist, etc.}, which can be seen from the full category results in appendix. We hypothesize that the failure is caused by the extreme sparsity of minor classes and the lack of discriminative raw features such as RGB colors.

\subsection{Ablation Study}
To evaluate the effectiveness of each component of our pipeline and the choices of hyperparameters, we conduct the following ablation experiments on Area-5 of S3DIS \cite{Armeni2017}. 

\textbf{(1) Only removing superpoint constructor}. In this setting, no superpoint will be constructed and there is no progressive growing mechanism as well. All per-point features will be directly clustered into 300 semantic primitives in training. Other settings are the same as the full method. 

\textbf{(2) Only removing semantic primitive clustering}. In this setting, the semantic primitive number $S$ is directly set as 12 (the actual number of classes). In both training and testing, all superpoints will be grouped as 12 categories. 

\textbf{(3) Only removing PFH feature in semantic primitive clustering}. This aims to evaluate the benefits of explicitly including geometry features for semantic primitive clustering. All other settings are kept the same.

\textbf{(4)$\sim$(6) Sensitivity to different voxel sizes of initial superpoints}. When constructing initial superpoints for each point cloud, here we select 3 sizes of voxel grids for VCCS algorithm, \ie{} \{25cm, 50cm, 75cm\}. Note that, in all experiments of our full method, we constantly set voxel size as 50cm. Intuitively, the smaller grid size, the more homogeneous initial superpoints, and the less semantics reserved. 

\textbf{(7)$\sim$(11) Sensitivity to different choices of $M^1$}. Since $M^0$ is determined by VCCS and we empirically find that the average number of initial superpoints across S3DIS is about 160, \ie{} $M^0\approx 160$, here we select 4 groups for $M^1$, \ie{} \{160, 120, 80, 40\}. In our full method, we always set $M^1$ as 80 (Algorithm \ref{alg:training}). 

\textbf{(12)$\sim$(15) Sensitivity to different choices of $M^T$}. $M^T$ is the ending value of $M$ during progressive growing, we conduct 4 ablations for $M^T$, \ie{} \{60, 40, 20, 12\}. In our full method, we always set $M^T$ as 20 (Algorithm \ref{alg:training}). 

\textbf{(16)$\sim$(19) Sensitivity to decreasing speeds}. We conduct 4 ablations for the decreasing speed of superpoint numbers in the growing process. After every \{40, 13, 5, 3\} epochs, the number of superpoints $M^t$ is reduced by 1 in training. In our full method, we set it as 13 (Algorithm \ref{alg:training}).

\textbf{(20)$\sim$(23) Sensitivity to different choices of $S$}. The choice of semantic primitive number $S$ also controls how aggressive our method aims to learn semantics. Here we select 4 different values for comparison, \ie{} \{100, 200, 300, 500\}. Our full method always sets $S$ as 300 (Algorithm \ref{alg:training}). 

\begin{figure}[t]
\setlength{\abovecaptionskip}{ 2 pt}
\setlength{\belowcaptionskip}{ -10 pt}
\centering
   \includegraphics[width=0.8\linewidth]{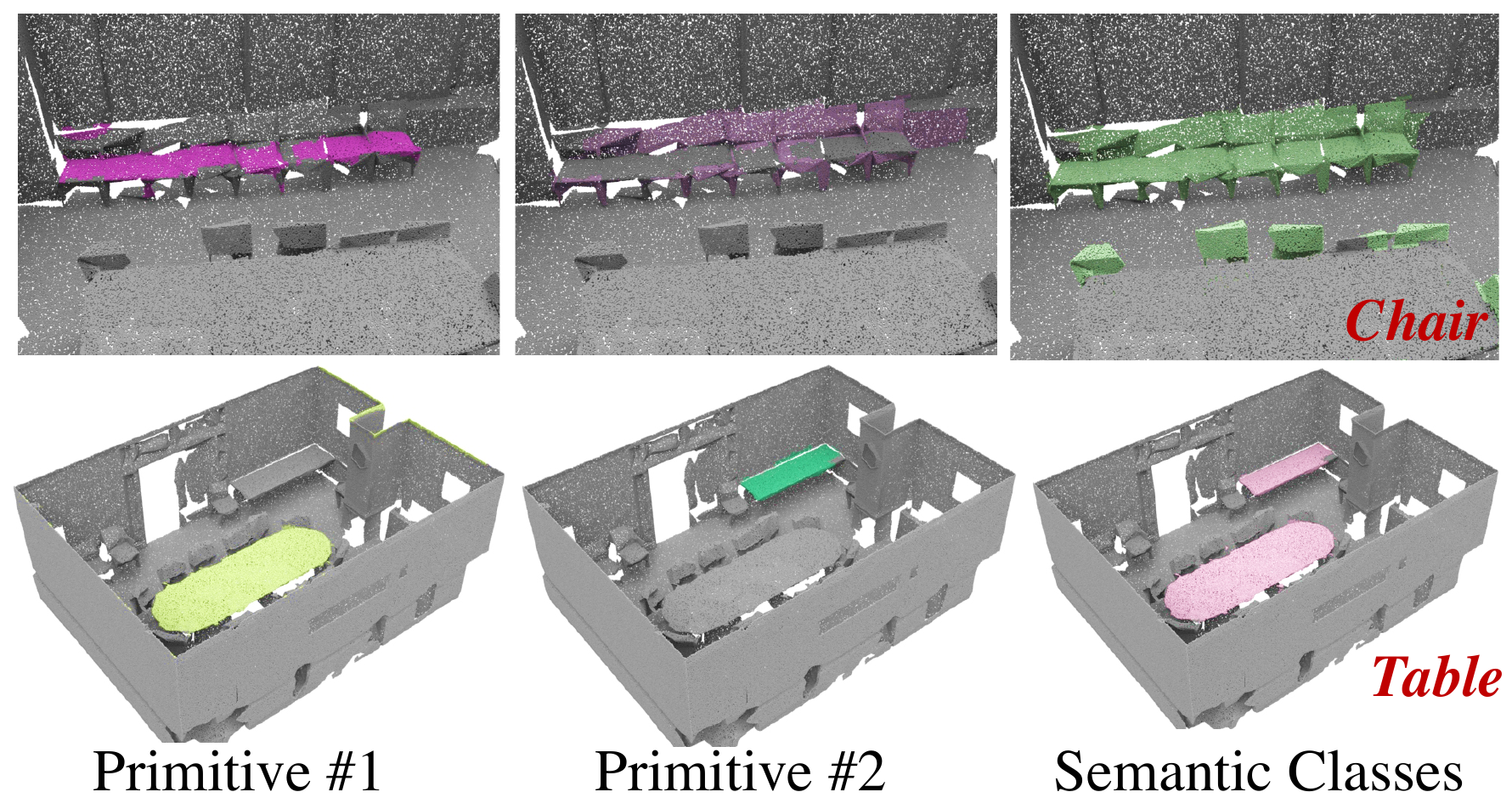}
\caption{Qualitative results of learned semantic primitives for two classes in S3DIS: chair/table.}\vspace{-0.3cm}
\label{fig:exp_ablation_primi}
\end{figure}

\textbf{Analysis:} As shown in Table \ref{tab:ablative}, we can see that: 
1) Without superpoint constructor, the method nearly fails with an mIoU score of just 20\%, unarguably demonstrating the key role it plays in automatically discovering complex semantic classes. 
2) The simple semantic primitive clustering module is also critical, if not the same important as superpoint constructor, showing that learning semantic elements instead of classes can effectively alleviate the difficulty. 3) Explicitly adding geometry based PFH features can indeed aid the network to get better results. 
4) Our method is extremely robust to different choices of all hyperparameters such as $M^1, M^T, S$, the voxel size for initial superpoints, and the decreasing speed. 
Figure \ref{fig:exp_ablation_primi} shows examples of learned semantic primitives. From this ablation study, we clearly see how the proposed components complement each other to achieve excellent performance without any human labels. 

\begin{table}[tb]
\centering
\caption{The mIoU scores of all ablated networks on Area-5 of S3DIS based on our full \nickname{}.}
\vspace{-0.3cm}
\label{tab:ablative}
\resizebox{0.4\textwidth}{!}{%
\begin{tabular}{lc}
\toprule[1.0pt]
 & mIoU(\%) \\
\toprule[1.0pt]
(1) Remove Superpoint Constructor &20.3$\pm$0.4 \\
(2) Remove Semantic Primitive Clustering &25.4$\pm$1.0 \\
(3) Remove PFH feature &38.9$\pm$0.9 \\
\toprule[1.0pt]
(4) 25cm voxels for initial superpoints &41.3$\pm$1.8 \\
\textbf{(5) 50cm voxels for initial superpoints} &\textbf{44.5}$\pm$1.1 \\
(6) 75cm voxels for initial superpoints &43.2$\pm$0.7 \\
\toprule[1.0pt]
(7) $M^1=160$ for progressive growing &43.3$\pm$1.3 \\
(8) $M^1=120$ for progressive growing &41.3$\pm$3.2 \\
\textbf{(9) $M^1=\textbf{80}$ for progressive growing} &\textbf{44.5}$\pm$1.1 \\
(10) $M^1=40$ for progressive growing &43.1$\pm$2.0 \\
\toprule[1.0pt]
(12) $M^T=60$ for progressive growing &42.4$\pm$1.1 \\
(13) $M^T=40$ for progressive growing &43.0$\pm$0.8 \\
\textbf{(14) $M^T=\textbf{20}$ for progressive growing} &\textbf{44.5}$\pm$1.1 \\
(15) $M^T=12$ for progressive growing &38.9$\pm$3.0 \\
\toprule[1.0pt]
(16) Decreasing speed 40 for progressive growing &42.2$\pm$2.4 \\
(17) Decreasing speed \textbf{13} progressive growing &\textbf{44.5}$\pm$1.1 \\
(18) Decreasing speed 5 for progressive growing &43.2$\pm$0.9 \\
(19) Decreasing speed 3 progressive growing &42.2$\pm$1.0 \\
\toprule[1.0pt]
(20) $S=100$ for semantic primitive clustering &41.4$\pm$1.2 \\
(21) $S=200$ for semantic primitive clustering &43.5$\pm$0.9 \\
\textbf{(22) $S=\textbf{300}$ for semantic primitive clustering} &\textbf{44.5}$\pm$1.1 \\
(23) $S=500$ for semantic primitive clustering &43.8$\pm$1.4 \\
\textbf{(24) The Full framework (\nickname{})} &\textbf{44.5}$\pm$1.1 \\
\toprule[1.0pt]
\end{tabular}%
}\vspace{-0.6cm}
\end{table}

\subsection{Comparison with Self-supervised Learning}
\label{sec:exp_ssl}


We further compare with existing self-supervised pre-training methods PointConstrast \cite{PC} (PC) and Contrastive Scene Contexts \cite{CSC} (CSC) in the following two groups of experiments. Note that, PC has two versions: InfoNCE and Hardest Contrastive, denoted as PC-I/PC-H. 
\vspace{-0.2cm}
\begin{itemize}[leftmargin=*]
\setlength{\itemsep}{1pt}
\setlength{\parsep}{1pt}
\setlength{\parskip}{1pt}
\item \textbf{Group 1:} All PC-I/PC-H/CSC/GrowSP are well-trained on the ScanNet training set. In fact, we simply reuse the official pre-trained models of PC-I/PC-H/CSC released by authors. All these trained networks are frozen, followed by K-means to group output features into semantic categories on both ScanNet (val set) and S3DIS (Area 5).
\item \textbf{Group 2:} We add a single linear classifier to each of the four pre-trained models, and only train the classifiers using full supervision until convergence (about 150 epochs for each method) on both ScanNet (training set) and S3DIS (Areas 1/2/3/4/6). After that, all four models are directly tested on both ScanNet (val set) and S3DIS (Area 5). Basically, this is to evaluate the quality of learned point features via linear probing.
\end{itemize}

\begin{table}[hb]
\begin{center}
\setlength{\belowcaptionskip}{ -10 pt}
\vspace{-0.2cm}
\caption{OA / mAcc / mIoU (\%) scores of Groups 1\&2.}
\vspace{-0.3cm}
\label{tab:exp_self-sup}
\resizebox{0.45\textwidth}{!}{
\begin{tabular}{lccccc}
\toprule[1.0pt]
\multirow{2}{*}{}  &\multicolumn{2}{c}{\textbf{Group 1} (K-means)} &\multicolumn{2}{c}{\textbf{Group 2} (Linear Probing)}\\
\toprule[1.0pt]
&ScanNet  &S3DIS &ScanNet &S3DIS \\
\toprule[1.0pt]
 PC-I \cite{PC} & 27.6 / 10.1 / 5.1 & 43.8 / 18.6 / 10.4 & 57.1 / 19.6 / 13.3 & 64.3 / 32.6 / 23.1 \\
 PC-H \cite{PC} & 29.5 / 12.5 / 5.8 & 42.8 / 17.5 / 11.3 & 62.6 / 18.8 / 13.3 & 63.4 / 36.3 / 25.9 \\
 CSC \cite{CSC}  & 44.9 / 11.8 / 7.7 & 43.3 / 22.4 / 13.5 & 69.3 / 29.5 / 21.8 & 78.2 / 43.6 / 35.3 \\
\textbf{Ours} & \textbf{62.9} / \textbf{44.3} /  \textbf{27.7} & \textbf{56.4} / \textbf{43.1} / \textbf{28.6} & \textbf{73.5} / \textbf{42.6} / \textbf{31.6} & \textbf{80.1} / \textbf{55.4} / \textbf{44.7}  \\
\toprule[1.0pt]
\vspace{-0.3cm}
\end{tabular}}
\end{center}
\end{table}
\vspace{-0.85cm}

\textbf{Analysis:} As shown in Table \ref{tab:exp_self-sup}: 1) All existing self-supervised methods completely fail to estimate semantics using K-means, while our method is significantly better. 2) Given an additional layer for linear probing, our method also shows clearly better segmentation results. We can see that the existing self-supervised pre-trained features actually do not have rich semantic category information, while our method can directly learn semantics.

\section{Conclusion}

We demonstrate that multiple 3D semantic classes can be automatically discovered using a purely unsupervised method from real-world point clouds. 
By leveraging a simple progressive growing strategy to create larger and larger superpoints over time, our method can successfully learn meaningful semantic elements. Extensive experiments validate the effectiveness of our approach. \\

\noindent\textbf{Acknowledgements:} This work was supported in part by Shenzhen Science and Technology Innovation Commission (JCYJ20210324120603011), in part by Research Grants Council of Hong Kong (25207822 \& 15225522).

\clearpage
{\small
\bibliographystyle{ieee_fullname}
\bibliography{references}
}
\clearpage
\appendix

{\large{\noindent\textbf{Appendix}}}

\begin{table*}[t!] 
\centering
\caption{Quantitative results of our method and baselines on the Area-1 of S3DIS dataset.}
\label{tab:s3dis_a1}
\resizebox{\textwidth}{!}
{
\begin{tabular}{crcccccccccccccccc}
\toprule[1.0pt]
& & OA(\%) & mAcc(\%)& mIoU(\%) & ceil. & floor & wall & beam & col. & wind. & door & table & chair & sofa & book. & board. \\
\toprule[1.0pt]
\multirow{3}{*}{\makecell[c]{Supervised\\Methods} }& PointNet \cite{Qi2016} & 75.4 & 74.8 & 55.0 & 88.3 & 93.2 & 69.2 & 49.5 & 37.8 & 74.5 & 65.6 & 41.2 & 42.5 & 22.3 & 35.4 & 40.9\\

& PointNet++ \cite{Qi2017} & 76.1 & 77.9 & 58.2 & 90.5 & 94.4 & 65.7 & 38.2 & 31.9 & 61.5 & 66.0 & 45.3 & 60.4 & 41.2 & 45.8 & 57.4\\
& SparseConv \cite{Graham2018} & 89.0 & 79.5 & 72.5 & 93.6 & 95.6 & 76.1 & 65.9 & 60.9 & 60.0 & 74.2 & 81.9 & 85.4 & 69.2 & 73.4  & 33.5\\
\toprule[1.0pt]
\multirow{8}{*}{\makecell[c]{Unsupervised\\Methods} } & RandCNN
 &24.9$\pm$1.1 &20.3$\pm$0.68 &10.9$\pm$0.9 &27.8$\pm$8.2 &29.0$\pm$4.0 &16.6$\pm$2.0 &10.7$\pm$5.4 &4.7$\pm$1.3 &14.8$\pm$6.4 &8.1$\pm$2.5 &4.1$\pm$0.8 &3.7$\pm$2.0 &0.2$\pm$0 &6.4$\pm$1.4 &5.1$\pm$2.9\\

& van Kmeans &20.9$\pm$1.1  &24.1$\pm$2.1 &10.1$\pm$0.9 &15.4$\pm$2.3 &17.8$\pm$2.3 &10.5$\pm$1.7 &16.8$\pm$0.3 &1.9$\pm$1.4 &16.0$\pm$4.1 &12.1$\pm$3.9 &9.9$\pm$1.6 &8.1$\pm$0.2 &0.1$\pm$0.2 &6.2$\pm$1.7 &6.7$\pm$2.2\\

& van Kmeans-S &21.8$\pm$0.9  &25.6$\pm$2.7 &10.8$\pm$0.6 &18.1$\pm$0.5 &20.4$\pm$0.2 &9.5$\pm$0.6 &14.5$\pm$3.7 &1.0$\pm$0.5 &17.2$\pm$3.5 &16.2$\pm$4.7 &9.6$\pm$1.4 &6.8$\pm$1.0 &0.6$\pm$1.1 &8.2$\pm$1.3 &7.6$\pm$0.8\\

& van Kmeans-PFH &26.4$\pm$0  &25.7$\pm$0.1 &12.6$\pm$0 &34.0$\pm$0.4 &27.4$\pm$0.1 &11.6$\pm$0.3 &17.0$\pm$0.1 &5.2$\pm$0.1 &15.4$\pm$0.6 &9.7$\pm$0.8 &10.8$\pm$0 &5.8$\pm$0.1 &1.6$\pm$0 &8.1$\pm$0.1 &4.3$\pm$0.8\\

& van Kmeans-S-PFH &25.1$\pm$0.6  &23.0$\pm$1.8 &10.8$\pm$0.7 &26.1$\pm$7.0 &18.5$\pm$5.3 &15.0$\pm$1.4 &4.6$\pm$7.0 &2.6$\pm$2.0 &20.5$\pm$3.9 &15.9$\pm$2.3 &10.4$\pm$1.4 &6.0$\pm$0.5 &0.6$\pm$0.7 &5.9$\pm$3.4 &3.7$\pm$2.4\\

& IIC \cite{Ji2019} &29.2$\pm$2.1  &14.3$\pm$0.4 &8.0$\pm$0.8 &17.0$\pm$2.9 &31.4$\pm$8.0 &25.6$\pm$0.1 &4.3$\pm$0.1 &11.1$\pm$0.4 &0$\pm$0 &2.6$\pm$0.8 &1.4$\pm$0.2 &0.7$\pm$0.6 &0$\pm$0 &0.2$\pm$0.1 &1.4$\pm$2.1\\

& IIC-S \cite{Ji2019} &31.3$\pm$0.2  &21.1$\pm$0.2 &8.2$\pm$0.1 &35.5$\pm$0.3 &11.6$\pm$1.0 &13.9$\pm$0.2 &1.3$\pm$0.4 &0.2$\pm$0.2 &13.0$\pm$0.4 &15.6$\pm$0.2 &5.1$\pm$0.1 &0.6$\pm$0 &0.3$\pm$0.6 &0.3$\pm$0 &7.6$\pm$0.1\\

& IIC-PFH \cite{Ji2019} &27.3$\pm$1.9  &12.0$\pm$1.1 &6.1$\pm$0.3 &17.1$\pm$2.4 &17.0$\pm$7.7 &23.4$\pm$4.9 &4.4$\pm$1.1 &0.3$\pm$0.2 &0.8$\pm$1.0 &4.0$\pm$3.7 &0.4$\pm$0.3 &0.4$\pm$0.4 &0$\pm$0 &4.9$\pm$5.7 &0.1$\pm$0.1\\

& IIC-S-PFH \cite{Ji2019} &24.1$\pm$0.3  &13.2$\pm$0.4 &7.2$\pm$0.2 &21.7$\pm$0.9 &16.8$\pm$3.8 &18.0$\pm$1.0 &5.0$\pm$0.6 &3.6$\pm$0.7 &2.9$\pm$1.7 &3.7$\pm$0.4 &4.4$\pm$0.4 &0.7$\pm$0.2 &0$\pm$0 &4.7$\pm$2.8 &4.5$\pm$1.6\\

& PICIE \cite{Cho2021}& 45.7$\pm$2.9  &28.3$\pm$2.4 &19.4$\pm$1.7 &77.2$\pm$5.4 &63.1$\pm$10.9 &24.5$\pm$2.4 &15.8$\pm$0.6 &3.3$\pm$2.8 &4.4$\pm$0.2 &9.6$\pm$0.6 &10.2$\pm$2.4 &14.7$\pm$0.6 &0$\pm$0 &9.9$\pm$0.8 &0$\pm$0\\

& PICIE-S \cite{Cho2021} &48.9$\pm$0.3  &30.3$\pm$1.6 &21.8$\pm$1.2 &77.66$\pm$8.4 &58.1$\pm$16.8 &39.0$\pm$2.1 &13.6$\pm$1.6 &2.5$\pm$2.2 &2.8$\pm$0.7 &11.8$\pm$0.7 &20.0$\pm$1.9 &17.1$\pm$1.9 &1.8$\pm$0.7 &17.1$\pm$2.3 &0.6$\pm$0.9\\

& PICIE-PFH \cite{Cho2021} &51.9$\pm$3.3  &38.4$\pm$2.3 &27.1$\pm$2.1 &59.8$\pm$1.5 &78.7$\pm$9.1 &37.5$\pm$5.5 &6.8$\pm$1.8 &9.5$\pm$1.3 &4.9$\pm$2.5 &26.4$\pm$2.0 &50.2$\pm$1.3 &29.5$\pm$3.0 &0$\pm$0 &22.8$\pm$1.3 &0.1$\pm$0.2\\

& PICIE-S-PFH \cite{Cho2021} &47.9$\pm$1.4  &38.9$\pm$3.0 &27.2$\pm$2.4 &71.5$\pm$1.2 &66.3$\pm$1.9 &23.9$\pm$2.3 &20.4$\pm$1.0 &8.8$\pm$3.4 &6.9$\pm$2.8 &12.9$\pm$3.3 &39.3$\pm$1.9 &32.2$\pm$1.6 &0.2$\pm$0.2 &42.0$\pm$1.2 &2.4$\pm$5.1\\

& \textbf{\nickname{}(Ours)} & \textbf{72.9}$\pm$0.8 & \textbf{60.4}$\pm$4.2 &\textbf{45.6}$\pm$1.4 &\textbf{94.2}$\pm$0.2 &\textbf{90.8}$\pm$0.6 &\textbf{52.7}$\pm$1.6 &\textbf{36.7}$\pm$5.8 &\textbf{19.7}$\pm$1.6 &\textbf{33.3}$\pm$4.9 &\textbf{35.8}$\pm$1.3 &\textbf{66.5}$\pm$0.7 &\textbf{72.6}$\pm$4.4 &\textbf{13.1}$\pm$7.5 &\textbf{31.2}$\pm$5.6 &\textbf{16.7}$\pm$1.2\\
\bottomrule[1.0pt]
\end{tabular}
}
\end{table*}

\begin{table*}[h!] 
\centering
\caption{Quantitative results of our method and baselines on the Area-2 of S3DIS dataset.}
\label{tab:s3dis_a2}
\resizebox{\textwidth}{!}
{
\begin{tabular}{crcccccccccccccccc}
\toprule[1.0pt]
& & OA(\%) & mAcc(\%)& mIoU(\%) & ceil. & floor & wall & beam & col. & wind. & door & table & chair & sofa & book. & board. \\
\toprule[1.0pt]
\multirow{3}{*}{\makecell[c]{Supervised\\Methods} }& PointNet \cite{Qi2016} & 72.5 & 55.5 & 36.6 & 79.2 & 87.4 & 64.9 & 14.5 & 8.2 & 14.8 & 39.6 & 28.8 & 64.0 & 7.8 & 24.4 & 5.1\\
& PointNet++ \cite{Qi2017} & 72.1 & 62.3 & 39.9 & 85.8 & 69.6 & 71.2 & 24.9 & 27.5 & 32.5 & 43.6 & 27.4 & 51.3 & 6.0 & 26.8 & 12.4\\
& SparseConv \cite{Graham2018} & 87.9 & 69.5 & 57.3 & 89.5 & 93.8 & 77.0 & 29.1 & 32.5 & 65.5 & 45.7 & 67.9 & 88.8 & 34.9 & 54.5  & 8.15\\
\toprule[1.0pt]
\multirow{8}{*}{\makecell[c]{Unsupervised\\Methods} } & RandCNN
 &20.4$\pm$1.2 &17.0$\pm$2.1 &7.6$\pm$0.8 &16.8$\pm$2.4 &19.2$\pm$4.3 &12.8$\pm$0.2 &1.6$\pm$0.3 &1.1$\pm$0.4 &1.5$\pm$2.1 &9.4$\pm$1.7 &2.9$\pm$0.3 &16.1$\pm$2.9 &0.3$\pm$0.4 &6.4$\pm$0.7 &\textbf{13.0}$\pm$1.0\\

& van Kmeans &17.6$\pm$0.6  &16.6$\pm$0.6 &6.4$\pm$0.3 &16.4$\pm$1.5 &15.6$\pm$2.3 &11.3$\pm$0.9 &3.3$\pm$0.5 &0.9$\pm$0.1 &0.4$\pm$0.6 &6.8$\pm$00.8 &3.7$\pm$0.6 &11.0$\pm$0.3 &1.4$\pm$0.3 &4.6$\pm$0.7 &1.5$\pm$0.7\\

& van Kmeans-S &17.3$\pm$0.4  &17.6$\pm$0.7 &6.3$\pm$0.2 &14.7$\pm$1.4 &15.0$\pm$0.9 &12.1$\pm$0.3 &3.0$\pm$1.6 &0.9$\pm$0.3 &0$\pm$0 &7.0$\pm$0.2 &3.6$\pm$0.4 &11.4$\pm$0.4 &1.9$\pm$0.1 &4.3$\pm$0.8 &2.0$\pm$1.6\\

& van Kmeans-PFH &21.8$\pm$0.4  &22.3$\pm$0.3 &9.0$\pm$0 &31.0$\pm$0.4 &21.1$\pm$1.5 &10.1$\pm$0.2 &3.0$\pm$0.1 &2.6$\pm$0 &\textbf{3.2}$\pm$1.0 &10.7$\pm$0.1 &5.3$\pm$1.8 &9.8$\pm$0.2 &4.0$\pm$0.2 &7.2$\pm$0.4 &0.6$\pm$0.9\\

& van Kmeans-S-PFH &21.5$\pm$1.5  &18.0$\pm$2.1 &8.0$\pm$0.3 &23.7$\pm$3.0 &21.1$\pm$3.3 &14.1$\pm$1.1 &3.8$\pm$0.9 &1.8$\pm$0.2 &1.8$\pm$1.6 &5.4$\pm$2.5 &4.3$\pm$1.5 &10.2$\pm$0.6 &2.5$\pm$0.2 &6.5$\pm$1.4 &0.9$\pm$0.8\\

& IIC \cite{Ji2019} &41.6$\pm$0.3  &16.8$\pm$0.3 &10.6$\pm$0.1 &33.0$\pm$0.6 &43.7$\pm$3.2 &27.6$\pm$0.7 &1.7$\pm$0.7 &0$\pm$0 &0$\pm$0 &5.6$\pm$1.0 &0.1$\pm$0.1 &13.0$\pm$2.2 &0$\pm$0 &2.8$\pm$0.1 &0$\pm$0\\

& IIC-S \cite{Ji2019} &30.3$\pm$0  &11.9$\pm$0.1 &6.2$\pm$0.1 &29.3$\pm$0.1 &7.6$\pm$0.2 &22.6$\pm$0.1 &0.5$\pm$0.5 &1.5$\pm$0 &0$\pm$0 &3.5$\pm$0 &0.3$\pm$0.2 &2.2$\pm$0.2 &1.8$\pm$1.6 &3.4$\pm$0.1 &1.1$\pm$0\\

& IIC-PFH \cite{Ji2019} &27.3$\pm$3.0  &10.5$\pm$0.2 &5.8$\pm$0.3 &16.0$\pm$0.6 &13.9$\pm$3.9 &24.9$\pm$2.6 &0.3$\pm$0.2 &0.1$\pm$0.1 &0$\pm$0 &4.0$\pm$0.3 &0.6$\pm$0.4 &8.3$\pm$4.2 &0.1$\pm$0.1 &0.8$\pm$0.1 &0.1$\pm$0.1\\

& IIC-S-PFH \cite{Ji2019} &26.8$\pm$0.1  &12.8$\pm$0.3 &6.8$\pm$0.1 &30.5$\pm$0.4 &12.5$\pm$0.7 &16.5$\pm$0.9 &1.1$\pm$0.1 &0.4$\pm$0.2 &1.1$\pm$0.1 &7.0$\pm$0.1 &1.1$\pm$0.1 &5.7$\pm$0.4 &0.4$\pm$0.5 &4.0$\pm$0.3 &1.2$\pm$0.1\\

& PICIE \cite{Cho2021} &48.3$\pm$2.3  &27.2$\pm$1.5 &17.4$\pm$0.6 &72.4$\pm$1.5 &44.2$\pm$6.9 &39.6$\pm$1.6 &6.2$\pm$1.2 &1.7$\pm$0 &0.5$\pm$0 &7.7$\pm$0.3 &4.1$\pm$0.9 &20.1$\pm$4.1 &0$\pm$0 &7.7$\pm$0.2 &3.6$\pm$0.4\\

& PICIE-S \cite{Cho2021} &55.5$\pm$0.9  &30.4$\pm$1.0 &21.5$\pm$0.4 &69.1$\pm$1.2 &68.7$\pm$3.3 &46.8$\pm$2.0 &3.1$\pm$0.3 &1.1$\pm$0.9 &0.4$\pm$0.3 &11.6$\pm$1.1 &6.6$\pm$4.4 &38.2$\pm$2.9 &3.0$\pm$0.6 &7.7$\pm$0.7 &1.8$\pm$1.2\\

& PICIE-PFH \cite{Cho2021} &55.8$\pm$3.4  &37.2$\pm$1.9 &25.7$\pm$1.4 &51.7$\pm$6.0 &82.6$\pm$6.3 &41.8$\pm$4.8 &2.5$\pm$1.0 &5.3$\pm$0.3 &1.0$\pm$0.6 &13.7$\pm$1.4 &40.6$\pm$1.5 &55.0$\pm$11.7 &0$\pm$0 &12.5$\pm$1.3 &2.1$\pm$0.6\\

& PICIE-S-PFH \cite{Cho2021} &55.4$\pm$1.1  &39.2$\pm$0.8 &26.2$\pm$0.7 &65.8$\pm$6.1 &77.7$\pm$2.2 &29.3$\pm$4.3 &4.6$\pm$0.9 &2.7$\pm$0.8 &0.5$\pm$0.5 &12.4$\pm$0.8 &26.7$\pm$3.5 &62.2$\pm$7.0 &1.0$\pm$0.1 &28.0$\pm$1.0 &3.7$\pm$0.3\\

& \textbf{\nickname{}(Ours)} & \textbf{79.0}$\pm$0.8 &\textbf{51.8}$\pm$3.1 &\textbf{39.1}$\pm$1.6 &\textbf{85.7}$\pm$1.2 &\textbf{88.2}$\pm$1.5 &\textbf{67.0}$\pm$1.1 &\textbf{12.0}$\pm$2.0 &\textbf{24.8}$\pm$7.0 &0$\pm$0 &\textbf{24.2}$\pm$1.6 &\textbf{51.2}$\pm$5.6 &\textbf{77.1}$\pm$4.7 &\textbf{4.1}$\pm$5.7 &\textbf{24.5}$\pm$8.4 & 0.2$\pm$0.3\\
\bottomrule[1.0pt]
\end{tabular}
}
\end{table*}

\begin{table*}[h!] 
\centering
\caption{Quantitative results of our method and baselines on the Area-3 of S3DIS dataset.}
\label{tab:s3dis_a3}
\resizebox{\textwidth}{!}
{
\begin{tabular}{crcccccccccccccccc}
\toprule[1.0pt]
& & OA(\%) & mAcc(\%)& mIoU(\%) & ceil. & floor & wall & beam & col. & wind. & door & table & chair & sofa & book. & board. \\
\toprule[1.0pt]
\multirow{3}{*}{\makecell[c]{Supervised\\Methods} }& PointNet \cite{Qi2016} & 78.2 & 74.9 & 57.7 & 90.3 & 96.9 & 66.9 & 55.5 & 15.1 & 60.0 & 67.7 & 51.8 & 54.8 & 27.6 & 56.0 & 50.0\\
& PointNet++ \cite{Qi2017} & 79.8 & 85.9 & 65.8 & 91.4 & 98.0 & 68.5 & 50.1 & 15.2 & 74.8 & 74.7 & 63.2 & 70.1 & 53.6 & 54.0 & 76.5\\
& SparseConv \cite{Graham2018} & 91.3 & 86.8 & 78.6 & 93.1 & 96.2 & 80.4 & 74.7 & 63.3 & 77.2 & 69.5 & 80.1 & 85.5 & 89.5 & 80.1  & 52.2\\
\toprule[1.0pt]
\multirow{8}{*}{\makecell[c]{Unsupervised\\Methods} } & RandCNN
 &25.3$\pm$3.7 &21.7$\pm$1.9 &10.8$\pm$1.6 &25.5$\pm$9.6 &32.6$\pm$12.3 &17.2$\pm$0.5 &4.9$\pm$2.8 &3.3$\pm$1.0 &10.6$\pm$3.2 &9.1$\pm$3.1 &4.2$\pm$1.9 &2.9$\pm$2.0 &0.9$\pm$0.9 &9.3$\pm$3.1 &8.8$\pm$1.1\\

& van Kmeans &21.3$\pm$02  &22.1$\pm$0.9 &9.43$\pm$0.1 &20.2$\pm$2.2 &20.6$\pm$0.96 &13.3$\pm$0.6 &5.7$\pm$2.0 &1.3$\pm$1.8 &2.3$\pm$0.9 &14.1$\pm$1.7 &6.8$\pm$0.3 &6.8$\pm$0.9 &3.7$\pm$0.7 &9.7$\pm$0.7 &8.6$\pm$1.5\\

& van Kmeans-S &21.2$\pm$1.2  &21.9$\pm$2.3 &9.3$\pm$0.8 &19.8$\pm$0.9 &18.8$\pm$2.8 &14.4$\pm$0.6 &5.0$\pm$0.9 &1.6$\pm$1.4 &1.5$\pm$0.6 &14.4$\pm$2.0 &7.4$\pm$0.8 &6.6$\pm$0.6 &4.2$\pm$0.1 &10.3$\pm$0.2 &7.7$\pm$2.7\\

& van Kmeans-PFH &24.7$\pm$0.5  &27.3$\pm$1.8 &11.6$\pm$0.4 &33.6$\pm$1.3 &23.0$\pm$2.3 &11.0$\pm$1.1 &8.9$\pm$0.4 &3.5$\pm$0.3 &10.8$\pm$2.0 &12.0$\pm$2.4 &8.4$\pm$0.7 &6.9$\pm$1.1 &3.7$\pm$0.6 &12.6$\pm$0.2 &4.9$\pm$1.0\\

& van Kmeans-S-PFH &23.3$\pm$1.7  &21.7$\pm$2.9 &9.7$\pm$1.2 &24.7$\pm$4.0 &20.1$\pm$3.4 &14.1$\pm$1.2 &5.4$\pm$4.1 &2.3$\pm$1.2 &5.2$\pm$5.1 &12.3$\pm$1.3 &5.1$\pm$2.4 &7.3$\pm$0.2 &3.1$\pm$0.9 &9.3$\pm$3.2 &8.0$\pm$1.1\\

& IIC \cite{Ji2019} &32.1$\pm$1.4  &15.4$\pm$1.9 &8.4$\pm$0.7 &20.5$\pm$2.4 &25.5$\pm$6.9 &31.4$\pm$1.0 &1.0$\pm$0.9 &6.9$\pm$5.7 &0.2$\pm$0.4 &3.2$\pm$1.1 &1.6$\pm$0.2 &0.3$\pm$0.1 &0$\pm$0 &10.6$\pm$3.6 &0$\pm$0\\

& IIC-S \cite{Ji2019} &30.3$\pm$0.1  &14.8$\pm$0.1 &7.1$\pm$0 &33.9$\pm$0.2 &10.1$\pm$0 &16.1$\pm$0.3 &2.0$\pm$0 &0.4$\pm$0 &5.1$\pm$0 &8.8$\pm$0.3 &0.6$\pm$0 &0.3$\pm$0.5 &2.7$\pm$0.5 &2.1$\pm$0.5 &5.5$\pm$0.2\\

& IIC-PFH \cite{Ji2019} &34.5$\pm$4.5  &11.8$\pm$2.1 &6.3$\pm$1.8 &22.6$\pm$2.1 &13.0$\pm$2.0 &32.4$\pm$1.9 &0.1$\pm$0.1 &0$\pm$0 &0.1$\pm$0.2 &1.9$\pm$2.3 &0.3$\pm$0.2 &0.3$\pm$0.3 &1.0$\pm$0.6 &2.3$\pm$3.1 &1.9$\pm$0.4\\

& IIC-S-PFH \cite{Ji2019} &25.5$\pm$1.3  &12.8$\pm$0.5 &6.6$\pm$0.1 &26.1$\pm$1.3 &8.7$\pm$1.6 &18.8$\pm$2.4 &1.3$\pm$0.6 &2.9$\pm$1.7 &0.8$\pm$0.6 &5.3$\pm$1.9 &1.6$\pm$1.9 &4.8$\pm$2.5 &1.7$\pm$0.6 &5.3$\pm$2.0 &2.6$\pm$1.1\\

& PICIE \cite{Cho2021} &40.4$\pm$1.8  &29.2$\pm$0.9 &16.2$\pm$0.5 &50.5$\pm$3.7 &49.6$\pm$14.5 &33.7$\pm$1.7 &13.2$\pm$1.7 &3.0$\pm$0.3 &1.8$\pm$1.6 &6.5$\pm$0.99 &8.9$\pm$7.9 &7.5$\pm$1.1 &3.5$\pm$3.0 &16.2$\pm$1.5 &0.4$\pm$0.7\\

& PICIE-S \cite{Cho2021} &52.8$\pm$1.5  &32.3$\pm$0.2 &22.4$\pm$0.4 &79.4$\pm$2.2 &80.0$\pm$1.7 &38.6$\pm$2.9 &9.1$\pm$0.6 &2.3$\pm$0.7 &3.9$\pm$1.9 &10.7$\pm$1.7 &15.9$\pm$9.0 &11.0$\pm$9.2 &0$\pm$0 &17.6$\pm$1.4 &0.6$\pm$0.2\\

& PICIE-PFH \cite{Cho2021} &57.8$\pm$3.7  &36.7$\pm$1.0 &26.4$\pm$1.6 &60.5$\pm$1.5 &78.8$\pm$11.1 &53.9$\pm$3.5 &4.1$\pm$2.7 &4.8$\pm$1.5 &0.3$\pm$0.6 &16.0$\pm$0.9 &50.0$\pm$1.0 &25.3$\pm$15.0 &1.0$\pm$1.3 &26.4$\pm$4.5 &0.6$\pm$0.4\\

& PICIE-S-PFH \cite{Cho2021} &51.8$\pm$1.7  &43.3$\pm$2.2 &29.3$\pm$2.5 &85.0$\pm$0.9 &72.6$\pm$2.9 &23.7$\pm$1.6 &14.2$\pm$0.4 &4.0$\pm$0.5 &5.7$\pm$2.0 &12.7$\pm$2.5 &48.5$\pm$10.2 &32.2$\pm$1.9 &2.8$\pm$1.8 &48.3$\pm$4.2 &2.1$\pm$0.3\\

& \textbf{\nickname{}(Ours)}&\textbf{74.2}$\pm$0.8 &\textbf{68.4}$\pm$1.3 &\textbf{47.7}$\pm$1.2 &\textbf{92.9}$\pm$0.1 &\textbf{91.7}$\pm$0.1 &\textbf{48.3}$\pm$3.3 &\textbf{49.3}$\pm$5.9 &\textbf{15.8}$\pm$2.3 &\textbf{21.1}$\pm$2.0 &\textbf{38.7}$\pm$1.8 &\textbf{60.6}$\pm$1.4 &\textbf{66.5}$\pm$2.9 &\textbf{28.5}$\pm$5.4 &\textbf{59.2}$\pm$6.6 &0$\pm$0\\
\bottomrule[1.0pt]
\end{tabular}
}
\end{table*}

\begin{table*}[h!] 
\centering
\caption{Quantitative results of our method and baselines on the Area-4 of S3DIS dataset.}
\label{tab:s3dis_a4}
\resizebox{\textwidth}{!}
{
\begin{tabular}{crcccccccccccccccc}
\toprule[1.0pt]
& & OA(\%) & mAcc(\%)& mIoU(\%) & ceil. & floor & wall & beam & col. & wind. & door & table & chair & sofa & book. & board. \\
\toprule[1.0pt]
\multirow{3}{*}{\makecell[c]{Supervised\\Methods} }& PointNet \cite{Qi2016} & 73.0 & 58.6 & 41.6 & 81.3 & 95.7 & 68.4 & 1.3 & 22.4 & 29.0 & 44.8 & 39.3 & 42.5 & 17.6 & 36.6 & 20.1\\
& PointNet++ \cite{Qi2017} & 74.8 & 66.4 & 47.7 & 85.5 & 96.1 & 68.9 & 4.4 & 23.8 & 27.0 & 50.5 & 44.9 & 54.0 & 35.6 & 38.4 & 43.8\\
& SparseConv \cite{Graham2018} & 88.3 & 76.2 & 65.5 & 93.0 & 94.9 & 78.2 & 53.3 & 57.9 & 43.4 & 59.1 & 69.4 & 76.6 & 55.1 & 73.8  & 30.9\\
\toprule[1.0pt]
\multirow{8}{*}{\makecell[c]{Unsupervised\\Methods} } & RandCNN
 &22.6$\pm$1.9 &18.3$\pm$2.0 &9.0$\pm$1.1 &22.0$\pm$9.7 &28.2$\pm$2.8 &16.4$\pm$1.7 &0.4$\pm$0 3 &4.2$\pm$0.5 &5.6$\pm$0.9 &11.9$\pm$0.7 &4.5$\pm$1.3 &4.3$\pm$0.5 &1.4$\pm$0.8 &6.4$\pm$0.9 &\textbf{2.3}$\pm$0.9\\

& van Kmeans &17.9$\pm$0.5  &19.9$\pm$0.5 &7.8$\pm$0.2 &18.6$\pm$1.2 &18.2$\pm$0.2 &10.6$\pm$1.0 &0.9$\pm$0.3 &3.8$\pm$1.6 &5.2$\pm$1.3 &11.7$\pm$1.3 &5.8$\pm$0.3 &7.4$\pm$1.8 &2.4$\pm$0.7 &8.7$\pm$1.0 &0.4$\pm$0.6\\

& van Kmeans-S &17.3$\pm$0.5  &19.6$\pm$1.0 &7.4$\pm$0.3 &16.7$\pm$0.5 &14.8$\pm$1.1 &10.2$\pm$1.0 &0.9$\pm$0.3 &4.4$\pm$1.8 &5.1$\pm$0.8 &10.5$\pm$0.8 &6.1$\pm$0.4 &7.5$\pm$1.4 &2.8$\pm$0.2 &9.8$\pm$0.3 &0$\pm$0\\

& van Kmeans-PFH &21.7$\pm$0.5  &22.3$\pm$0.8 &9.3$\pm$0.3 &28.9$\pm$1.0 &21.3$\pm$0.9 &12.6$\pm$0.7 &1.2$\pm$0.1 &5.9$\pm$1.0 &4.4$\pm$0.8 &11.1$\pm$0.6 &6.6$\pm$0.5 &8.6$\pm$0 &2.6$\pm$0.7 &7.4$\pm$0.7 &0.7$\pm$0.2\\

& van Kmeans-S-PFH &23.1$\pm$3.6  &18.3$\pm$2.3 &8.9$\pm$1.3 &24.4$\pm$6.5 &26.2$\pm$7.6 &14.5$\pm$2.5 &1.7$\pm$1.0 &5.0$\pm$1.1 &3.2$\pm$2.0 &11.7$\pm$0.8 &3.8$\pm$0.8 &6.1$\pm$1.3 &0.4$\pm$0.7 &9.0$\pm$0.4 &1.1$\pm$0.3\\

& IIC \cite{Ji2019} &33.0$\pm$0.3  &13.5$\pm$0.5 &8.2$\pm$0.2 &14.9$\pm$1.0 &25.9$\pm$3.9 &35.1$\pm$0.4 &0$\pm$0 &1.1$\pm$0.4 &1.4$\pm$1.4 &3.7$\pm$1.0 &4.1$\pm$2.0 &0.9$\pm$0.1 &0$\pm$0 &10.8$\pm$0.7 & 0$\pm$0\\

& IIC-S \cite{Ji2019} &26.7$\pm$0.2  &13.5$\pm$0.1 &6.1$\pm$0.1 &27.6$\pm$0 &4.9$\pm$0.9 &15.3$\pm$0 &0.1$\pm$0.1 &1.0$\pm$0.1 &5.3$\pm$0.1 &12.0$\pm$0 &1.5$\pm$1.9 &2.3$\pm$1.9 &1.1$\pm$0.9 &1.6$\pm$0 &0$\pm$0\\

& IIC-PFH \cite{Ji2019} &31.0$\pm$7.6  &11.4$\pm$1.5 &6.3$\pm$1.4 &21.0$\pm$1.5 &10.9$\pm$2.1 &30.0$\pm$6.2 &0.1$\pm$0.2 &1.3$\pm$1.4 &3.0$\pm$4.0 &2.3$\pm$0.4 &0.8$\pm$0.1 &1.3$\pm$1.2 &0$\pm$0 &4.6$\pm$4.1 &0.1$\pm$0.1\\

& IIC-S-PFH \cite{Ji2019} &27.3$\pm$2.0  &13.2$\pm$0.4 &6.9$\pm$0.1 &29.4$\pm$1.9 &9.9$\pm$0.4 &20.0$\pm$4.9 &0.1$\pm$0 &2.6$\pm$1.5 &3.1$\pm$1.2 &4.0$\pm$0.9 &3.7$\pm$1.1 &2.5$\pm$1.1 &0.6$\pm$0.1 &6.3$\pm$1.9 &0.8$\pm$0.5 \\

& PICIE \cite{Cho2021} &43.2$\pm$0.6  &29.4$\pm$1.2 &17.8$\pm$0.5 &62.2$\pm$1.3 &72.7$\pm$2.1 &22.6$\pm$0.7 &2.5$\pm$0.6 &3.4$\pm$0.2 &3.5$\pm$0.4 &8.8$\pm$1.1 &4.1$\pm$7.0 &17.4$\pm$1.1 &0$\pm$0 &15.5$\pm$1.7 &0.7$\pm$0.1\\

& PICIE-S \cite{Cho2021} &50.0$\pm$1.2  &32.2$\pm$1.1 &22.3$\pm$0.4 &72.7$\pm$2.1 &77.7$\pm$2.8 &37.4$\pm$1.6 &1.6$\pm$1.1 &2.4$\pm$0.7 &3.3$\pm$1.3 &11.3$\pm$1.1 &25.8$\pm$1.4 &19.7$\pm$1.5 &3.5$\pm$1.5 &12.3$\pm$3.5 &0$\pm$0\\

& PICIE-PFH \cite{Cho2021} &58.6$\pm$2.6  &42.6$\pm$4.2 &28.5$\pm$1.3 &69.5$\pm$6.8 &83.6$\pm$7.3 &48.3$\pm$3.2 &2.6$\pm$1.6 &9.4$\pm$6.4 &5.8$\pm$1.0 &16.4$\pm$1.1 &44.1$\pm$0.8 &36.3$\pm$3.4 &1.6$\pm$0.6 &23.9$\pm$3.3 &0.1$\pm$0.2\\

& PICIE-S-PFH \cite{Cho2021} &47.1$\pm$1.5  &44.8$\pm$0.8 &27.6$\pm$0.5 &67.2$\pm$3.5 &67.8$\pm$3.5 &25.9$\pm$2.2 &2.6$\pm$0.2 &14.5$\pm$0.5 &5.2$\pm$0.8 &10.2$\pm$0.2 &47.9$\pm$1.1 &44.1$\pm$1.0 &0.9$\pm$0.2 &44.2$\pm$0.7 &0.4$\pm$0.7\\

& \textbf{\nickname{}(Ours)} &\textbf{76.0}$\pm$0.9 &\textbf{59.8}$\pm$1.0 &\textbf{42.8}$\pm$1.1 &\textbf{90.6}$\pm$0.7 &\textbf{91.5}$\pm$0.7 &\textbf{64.4}$\pm$1.1 &\textbf{15.9}$\pm$5.3 &\textbf{7.60}$\pm$3.3 &\textbf{27.4}$\pm$1.9 &\textbf{31.5}$\pm$0.6 &\textbf{52.0}$\pm$1.2 &\textbf{67.4}$\pm$1.6 &\textbf{16.8}$\pm$6.6 &\textbf{48.5}$\pm$5.1 &0$\pm$0\\
\bottomrule[1.0pt]
\end{tabular}
}
\end{table*}

\begin{table*}[h!] 
\centering
\caption{Quantitative results of our method and baselines on the Area-5 of S3DIS dataset.}
\label{tab:s3dis_a5}
\resizebox{\textwidth}{!}
{
\begin{tabular}{crcccccccccccccccc}
\toprule[1.0pt]
& & OA(\%) & mAcc(\%)& mIoU(\%) & ceil. & floor & wall & beam & col. & wind. & door & table & chair & sofa & book. & board. \\
\toprule[1.0pt]
\multirow{3}{*}{\makecell[c]{Supervised\\Methods} }& PointNet \cite{Qi2016}& 77.5 & 59.1 & 44.6 & 85.2 & 97.4 & 72.3 & 0.1 & 10.6 & 54.9 & 18.5 & 48.4 & 39.5 & 12.4 & 55.5 & 40.2\\
& PointNet++ \cite{Qi2017}& 77.5 & 62.6 & 50.1 & 83.1 & 97.2 & 66.4 & 0 & 8.1 & 55.6 & 15.2 & 60.4 & 64.5 & 36.6 & 58.3 & 55.7\\
& SparseConv \cite{Graham2018} & 88.4 & 69.2 & 60.8 & 92.6 & 95.9 & 77.2 & 0.1 & 36.7 & 37.6 & 59.8 & 77.2 & 83.9 & 59.7 & 78.5  & 30.39\\
\toprule[1.0pt]
\multirow{8}{*}{\makecell[c]{Unsupervised\\Methods} } & RandCNN
 &23.3$\pm$2.6 &17.3$\pm$1.1 &9.2$\pm$1.2 &25.3$\pm$3.4 &24.5$\pm$3.4 &17.4$\pm$1.5 &0$\pm$0 &2.3$\pm$0.8 &12.5$\pm$2.4 &6.5$\pm$1.2 &5.7$\pm$1.7 &3.0$\pm$0.3 &0.3$\pm$0.2 &10.1$\pm$2.0 &2.2$\pm$0.9\\

& van Kmeans &21.4$\pm$0.6  &21.2$\pm$1.6 &8.7$\pm$0.3 &18.7$\pm$2.6 &18.0$\pm$1.2 &16.7$\pm$0.2 &0.2$\pm$0 &2.5$\pm$0.5 &12.0$\pm$02 &5.7$\pm$02.3 &8.7$\pm$0.6 &5.6$\pm$1.0 &0$\pm$0 &13.6$\pm$1.0 &2.3$\pm$1.3\\

& van Kmeans-S &21.9$\pm$0.5  &22.9$\pm$0.4 &9.0$\pm$0.2 &19.3$\pm$1.1 &18.1$\pm$0.7 &17.0$\pm$1.3 &0.2$\pm$1.3 &2.1$\pm$0.2 &11.8$\pm$0 &4.5$\pm$1.1 &8.9$\pm$0.4 &6.6$\pm$0.7 &0.2$\pm$0.4 &14.0$\pm$1.8 &4.8$\pm$0.3\\

& van Kmeans-PFH &23.2$\pm$0.7  &23.6$\pm$1.7 &10.2$\pm$1.4 &32.0$\pm$0.6 &20.5$\pm$1.5 &10.3$\pm$0.9 &0.1$\pm$0.1 &3.6$\pm$0.7 &15.2$\pm$1.3 &7.1$\pm$0.5 &9.9$\pm$0.2 &6.2$\pm$0.1 &0.7$\pm$0.1 &12.4$\pm$0.2 &\textbf{4.9}$\pm$0.1\\

& van Kmeans-S-PFH &22.8$\pm$1.7  &20.6$\pm$0.7 &9.2$\pm$0.9 &25.2$\pm$3.9 &26.5$\pm$8.0 &12.7$\pm$0.7 &0.4$\pm$0.1 &2.0$\pm$0.7 &8.7$\pm$0.7 &8.1$\pm$3.6 &5.8$\pm$1.7 &5.7$\pm$1.0 &0$\pm$0 &12.7$\pm$1.8 &3.3$\pm$0.6\\

& IIC \cite{Ji2019} &28.5$\pm$0.2  &12.5$\pm$0.2 &6.4$\pm$0 &6.1$\pm$0.8 &19.8$\pm$0.7 &27.9$\pm$0.5 &0$\pm$0 &2.1$\pm$0.1 &0.1$\pm$0.1 &3.4$\pm$0.1 &7.9$\pm$0.2 &0.4$\pm$0.3 &0$\pm$0 &8.6$\pm$0.5 &0$\pm$0\\

& IIC-S \cite{Ji2019} &29.2$\pm$0.5  &13.0$\pm$0.2 &6.8$\pm$0 &28.9$\pm$0.7 &12.3$\pm$0.3 &18.7$\pm$0.2 &0$\pm$0 &0.1$\pm$0 &3.6$\pm$0.1 &1.3$\pm$0.2 &3.8$\pm$0.3 &0.6$\pm$0 &0$\pm$0 &8.1$\pm$0.1 &3.8$\pm$0\\

& IIC-PFH \cite{Ji2019} &28.6$\pm$0.1  &16.8$\pm$0.1 &7.9$\pm$0.4 &23.7$\pm$0.1 &24.9$\pm$0.1 &17.7$\pm$0.3 &\textbf{5.9}$\pm$0.1 &1.4$\pm$0.1 &12.6$\pm$0.1 &0.2$\pm$0 &5.3$\pm$0.1 &0.6$\pm$0 &0$\pm$0 &2.4$\pm$0.1 &0$\pm$0\\

& IIC-S-PFH \cite{Ji2019} &31.2$\pm$0.2  &16.3$\pm$0.1 &9.1$\pm$0.1 &43.2$\pm$0.1 &23.6$\pm$0.2 &14.9$\pm$0.5 &0$\pm$0 &1.6$\pm$0 &3.9$\pm$0.1 &2.4$\pm$0.1 &3.6$\pm$0 &1.5$\pm$0 &0.8$\pm$0 &9.5$\pm$0.2 &4.5$\pm$0.1\\

& PICIE \cite{Cho2021} &61.6$\pm$1.5  &25.8$\pm$1.6 &17.9$\pm$0.9 &65.7$\pm$7.4 &61.4$\pm$12.3 &58.4$\pm$0.4 &0$\pm$0 &0.3$\pm$0.4 &2.2$\pm$2.7 &1.7$\pm$1.2 &12.1$\pm$8.8 &0$\pm$0 &0$\pm$0 &12.4$\pm$2.0 &0$\pm$0\\

& PICIE-S \cite{Cho2021} &49.6$\pm$2.8  &28.9$\pm$1.0 &20.0$\pm$0.6 &64.2$\pm$5.8 &75.1$\pm$3.7 &42.4$\pm$2.4 &0.1$\pm$0 &1.2$\pm$0.1 &4.6$\pm$0.5 &7.4$\pm$2.6 &18.7$\pm$0.9 &9.2$\pm$4.6 &1.0$\pm$0.2 &16.0$\pm$2.2 &0.4$\pm$0.2\\

& PICIE-PFH \cite{Cho2021} &54.0$\pm$0.8  &36.8$\pm$1.7 &24.4$\pm$0.6 &58.4$\pm$5.0 &68.6$\pm$4.5 &49.9$\pm$2.0 &0.1$\pm$0.2 &7.6$\pm$0.7 &5.3$\pm$0.8 &14.2$\pm$1.8 &45.1$\pm$1.2 &16.3$\pm$1.0 &0.1$\pm$0 &27.1$\pm$5.6 &0.6$\pm$0.5\\

& PICIE-S-PFH \cite{Cho2021} &48.4$\pm$0.9  &40.4$\pm$1.6 &25.2$\pm$1.2 &59.6$\pm$2.4 &72.5$\pm$1.7 &26.0$\pm$1.3 &0.2$\pm$0 &8.5$\pm$3.2 &5.9$\pm$0.1 &8.7$\pm$0.7 &46.0$\pm$3.7 &26.9$\pm$4.6 &0.4$\pm$0.1 &46.8$\pm$4.1 &0.3$\pm$0\\

& \textbf{\nickname{}(Ours)} &\textbf{78.4}$\pm$1.5 &\textbf{57.2}$\pm$1.7 &\textbf{44.5}$\pm$1.1 &\textbf{90.45}$\pm$2.1 &\textbf{90.1}$\pm$2.1 &\textbf{66.7}$\pm$1.7 &0$\pm$0 &\textbf{14.8}$\pm$6.6 &\textbf{27.6}$\pm$3.6 &\textbf{45.6}$\pm$1.2 &\textbf{59.4}$\pm$1.6 &\textbf{71.9}$\pm$2.8 &\textbf{10.7}$\pm$4.0 &\textbf{56.0}$\pm$4.2 &0.2$\pm$0.1\\
\bottomrule[1.0pt]
\end{tabular}
}
\end{table*}

\begin{table*}[h!] 
\centering
\caption{Quantitative results of our method and baselines on the Area-6 of S3DIS dataset.}
\label{tab:s3dis_a6}
\resizebox{\textwidth}{!}
{
\begin{tabular}{crcccccccccccccccc}
\toprule[1.0pt]
& & OA(\%) & mAcc(\%)& mIoU(\%) & ceil. & floor & wall & beam & col. & wind. & door & table & chair & sofa & book. & board. \\
\toprule[1.0pt]
\multirow{3}{*}{\makecell[c]{Supervised\\Methods} }& PointNet \cite{Qi2016} & 79.0 & 79.6 & 60.9 & 85.7 & 96.5 & 71.8 & 59.4 & 47.4 & 67.4 & 74.3 & 56.2 & 48.9 & 20.9 & 50.0 & 52.5\\
& PointNet++ \cite{Qi2017} & 82.0 & 89.3 & 69.0 & 87.5 & 96.3 & 76.8 & 66.4 & 54.4 & 72.1 & 77.4 & 64.3 & 66.5 & 43.7 & 51.8 & 70.2\\
& SparseConv \cite{Graham2018} & 91.6 & 87.3 & 80.5 & 97.4 & 95.0 & 83.4 & 83.0 & 75.1 & 81.1 & 74.9 & 81.3 & 84.3 & 79.0 & 80.7  & 61.4\\
\toprule[1.0pt]
\multirow{8}{*}{\makecell[c]{Unsupervised\\Methods} } & RandCNN
 &22.1$\pm$2.6 &16.0$\pm$1.1 &8.5$\pm$1.2 &17.6$\pm$8.9 &24.2$\pm$3.3 &19.2$\pm$1.5 &0$\pm$0 &1.7$\pm$0.8 &12.2$\pm$2.4 &7.6$\pm$1.2 &6.2$\pm$1.7 &2.6$\pm$0.3 &0.2$\pm$0.2 &8.9$\pm$2.0 &1.7$\pm$8\\

& van Kmeans &21.0$\pm$0.2  &25.0$\pm$2.4 &10.4$\pm$1.0 &18.6$\pm$2.5 &17.6$\pm$2.4 &8.9$\pm$1.6 &11.3$\pm$5.4 &0.6$\pm$0.2 &14.8$\pm$0.8 &17.6$\pm$1.2 &12.0$\pm$0.7 &8.7$\pm$1.3 &0.3$\pm$0.6 &7.8$\pm$0.4 &\textbf{6.2}$\pm$1.2\\

& van Kmeans-S &20.6$\pm$1.5  &26.3$\pm$2.2 &10.2$\pm$0.8 &16.6$\pm$1.2 &17.8$\pm$2.5 &9.0$\pm$0.9 &13.4$\pm$6.4 &0.6$\pm$0.2 &15.0$\pm$0.8 &17.0$\pm$2.8 &10.7$\pm$2.5 &8.3$\pm$1.4 &1.8$\pm$0.8 &7.4$\pm$0.6 &5.1$\pm$1.3\\

& van Kmeans-PFH &25.8$\pm$0.2  &27.2$\pm$0.1 &12.8$\pm$0.1 &32.0$\pm$0.3 &25.1$\pm$0.9 &12.2$\pm$0 &15.0$\pm$0.1 &3.6$\pm$0.2 &19.7$\pm$1.0 &14.9$\pm$0.6 &9.4$\pm$0.1 &10.9$\pm$0 & 2.1$\pm$0 &5.3$\pm$0 &3.5$\pm$0.1\\

& van Kmeans-S-PFH &23.9$\pm$1.6  &23.0$\pm$2.4 &10.5$\pm$1.3 &22.2$\pm$5.6 &20.8$\pm$4.2 &15.2$\pm$2.8 &14.9$\pm$7.5 &0.3$\pm$0.3 &9.5$\pm$6.0 &13.6$\pm$1.9 &10.9$\pm$2.8 &7.4$\pm$3.6 &1.6$\pm$1.3 &5.0$\pm$1.3 &4.9$\pm$0.3\\

& IIC \cite{Ji2019} &32.5$\pm$2.1  &15.9$\pm$1.2 &9.2$\pm$0.9 &21.9$\pm$2.7 &33.8$\pm$6.9 &29.1$\pm$0.5 &3.1$\pm$0.2 &\textbf{15.2}$\pm$0.8 &0$\pm$0 &2.7$\pm$0.3 &0.74$\pm$0.3 &0$\pm$0 &0$\pm$0 &1.5$\pm$0.1 &1.8$\pm$1.6\\

& IIC-S \cite{Ji2019} &27.0$\pm$0.2  &16.5$\pm$0.2 &7.6$\pm$0.1 &28.9$\pm$0.4 &8.5$\pm$0.1 &12.7$\pm$0.5 &0.5$\pm$0.1 &0.2$\pm$0 &0$\pm$0 &14.1$\pm$0.1 &10.4$\pm$0.4 &0.6$\pm$0 &0$\pm$0 &0.96$\pm$0.1 &3.8$\pm$0\\

& IIC-PFH \cite{Ji2019} &28.4$\pm$0.3  &16.7$\pm$0.2 &7.8$\pm$0.1 &23.6$\pm$0.2 &24.5$\pm$0.2 &17.4$\pm$0.4 &5.8$\pm$0.2 &1.5$\pm$0 &12.6$\pm$0.1 &0.2$\pm$0 &4.8$\pm$0.2 &0.6$\pm$0.1 &0$\pm$0 &2.4$\pm$0.1 &0$\pm$0\\

& IIC-S-PFH \cite{Ji2019} &22.9$\pm$0.2  &13.1$\pm$0 &6.7$\pm$0.1 &28.8$\pm$0.2 &9.0$\pm$0.2 &12.6$\pm$0.1 &0.6$\pm$0 &4.8$\pm$0 &2.1$\pm$0.1 &6.6$\pm$0.1 &4.4$\pm$0.1 &1.2$\pm$0.1 &0.3$\pm$0 &5.4$\pm$0.1 &4.9$\pm$0.2 \\

& PICIE \cite{Cho2021} &39.3$\pm$4.8  &28.5$\pm$0.2 &17.8$\pm$1.5 &56.9$\pm$1.9 &61.7$\pm$17.5 &18.6$\pm$0.5 &20.5$\pm$1.9 &4.2$\pm$1.2 &6.0$\pm$0.4 &8.7$\pm$0 &14.7$\pm$1.7 &15.9$\pm$0.2 &1.1$\pm$1.8 &5.7$\pm$0.9 &0$\pm$0\\

& PICIE-S \cite{Cho2021} &47.4$\pm$0.9  &30.7$\pm$0.3 &21.8$\pm$0.2 &65.5$\pm$1.1 &67.5$\pm$3.4 &37.2$\pm$4.3 &16.6$\pm$1.9 &2.0$\pm$1.8 &4.8$\pm$1.2 &10.6$\pm$3.2 &23.7$\pm$1.5 &23.4$\pm$1.2 &23.4$\pm$1.5 &1.7$\pm$0.9 &0$\pm$0\\

& PICIE-PFH \cite{Cho2021} &51.8$\pm$2.6  &41.3$\pm$1.0 &27.9$\pm$0.9 &63.1$\pm$14.1 &56.5$\pm$6.5 &39.0$\pm$4.0 &11.4$\pm$3.5 &10.0$\pm$0.9 &5.3$\pm$0.4 &19.1$\pm$3.0 &63.8$\pm$0.6 &50.4$\pm$0.5 &0.5$\pm$0.9 &37.2$\pm$4.3 &0.9$\pm$1.5\\

& PICIE-S-PFH \cite{Cho2021} &44.0$\pm$0.5  &36.1$\pm$0.9 &24.7$\pm$0.4 &58.2$\pm$2.5 &60.4$\pm$1.2 &24.5$\pm$1.0 &17.9$\pm$3.0 &10.1$\pm$2.8 &8.1$\pm$1.1 &12.7$\pm$2.6 &44.0$\pm$0.2 &5.5$\pm$0.1 &0.5$\pm$0.9 &54.3$\pm$1.3 &0.3$\pm$0.6\\

& \textbf{\nickname{}(Ours)} &\textbf{75.6}$\pm$0.8 &\textbf{58.5}$\pm$1.3 &\textbf{47.6}$\pm$0.9 &\textbf{89.4}$\pm$0.3 &\textbf{88}$\pm$3.3 &\textbf{57.7}$\pm$0.5 &\textbf{70.6}$\pm$0.7 &2.0$\pm$2.2 &\textbf{32.4}$\pm$2.5 &\textbf{36.7}$\pm$1.5 &\textbf{63.2}$\pm$0.9 &\textbf{69.8}$\pm$0.8 &1.5$\pm$1.4 &\textbf{58.9}$\pm$11.8 &0.2$\pm$0.3\\
\bottomrule[1.0pt]
\end{tabular}
}
\end{table*}


\section{Details of Feature Extractor}
\label{sec:app_featExt}

\begin{figure}[ht]
\vspace{-0.3cm}
\setlength{\abovecaptionskip}{ 2 pt}
\setlength{\belowcaptionskip}{ -12 pt}
\centering
   \includegraphics[width=0.85\linewidth]{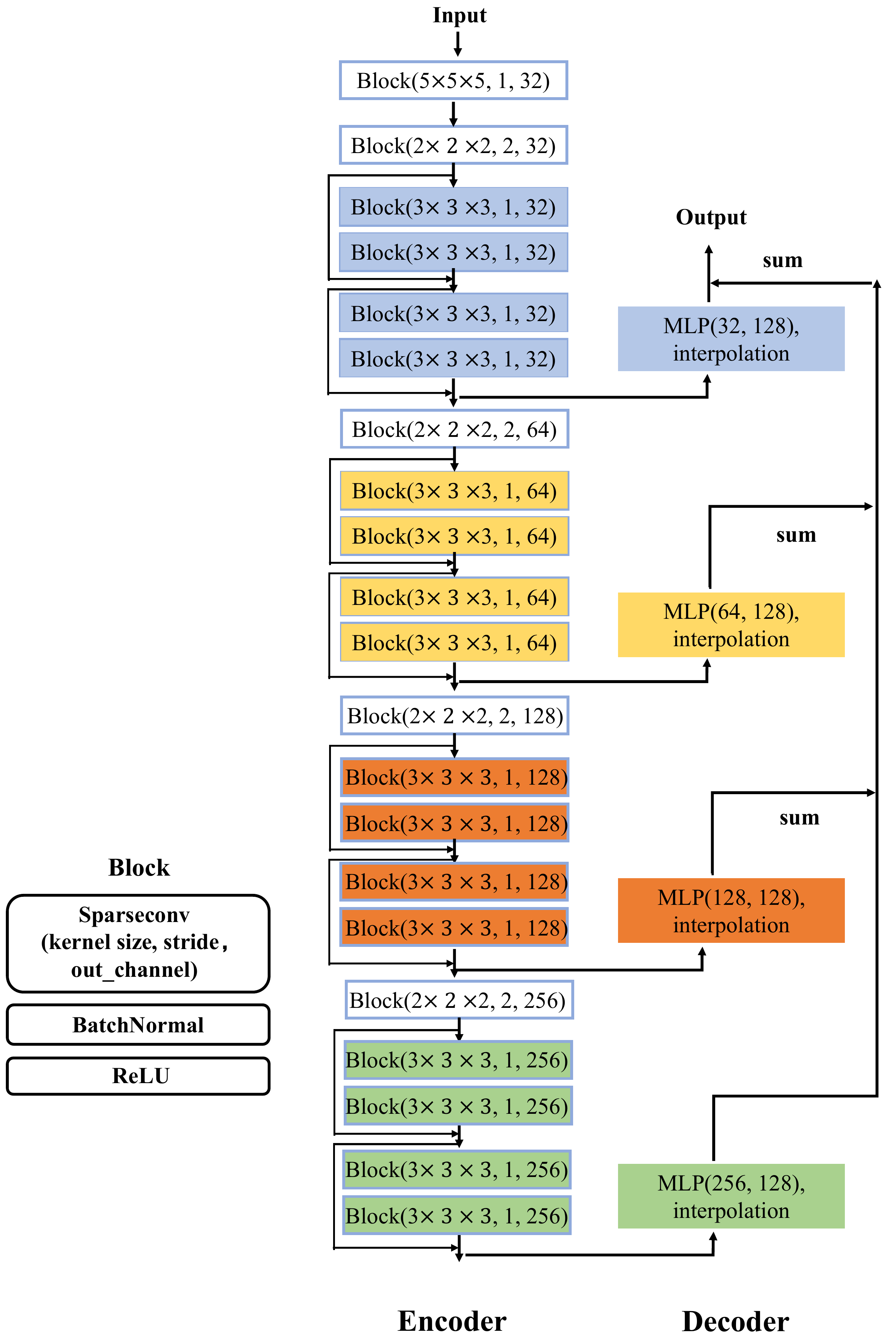}
\caption{Details of feature extractor (SparseConv \cite{Graham2018} with the Res16 architecture).}
\label{fig:app_architecture}
\end{figure}

The feature extractor of our \nickname{} simply follows the successful SparseConv\cite{Graham2018} architecture. Particularly, we use the existing implementation of MinkowskiEngine PyTorch package \cite{Choy2019}. 
As illustrated in \ref{fig:app_architecture}, the encoder uses Res16, and the decoder instead consists of 4 MLP layers which produce 128-dimensional features for interpolations to obtain the final per-point features.  

\section{Details of K-means}

In both progressive growing of superpoints and semantic primitive clustering, we adopt the simple K-means algorithm. Particularly, we use the existing package provided by Scikit-learn. \textit{\urlstyle{sf}\url{https://scikitlearn.org/stable/modules/generated/sklearn.cluster.KMeans.html}}

\section{Initial Superpoints Construction}
\label{sec:app_initSup}

As discussed in Section \ref{sec:meth_constructor} of the main paper,  initial superpoints are generated by VCCS \cite{Papon2013} combined with a Region Growing algorithm \cite{Adams1994}, 
which jointly consider the spatial/normal/normalized RGB information of 3D points to construct initial superpoints for indoor 3D point clouds.

\textbf{Details of VCCS}\cite{Papon2013}: VCCS incrementally expands superpoints from a set of seed points distributed evenly in 3D space on a voxel grid with resolution $R_{seed}$. 

\begin{figure}[ht]
\centering
   \includegraphics[width=1\linewidth]{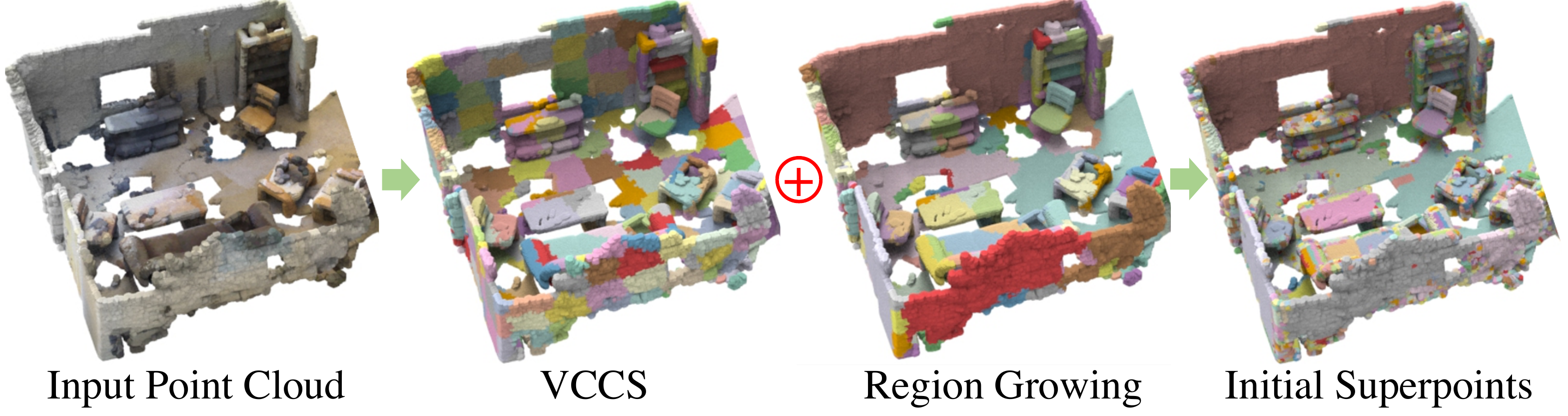}
\caption{\small{Example of VCCS, Region Growing, Initial Superpoints.}}
\label{fig:app_vccs_regionGrow}
\end{figure}

In our experiments of S3DIS and ScanNet datasets, we firstly voxelize  input point clouds into $5\times5\times5$ cm voxel grids. Secondly, a set of seed points are distributed evenly in the voxelized point clouds with an interval of 50cm. For each seed, in its 50cm radius sphere, we  set the seed point as an initial center and search its 27 neighbors, and compute a distance between each neighbor and the center as below:
\begin{equation}\label{eq:vccs}
    \boldsymbol{{D}} = \sqrt{w_{c}D^2_{c}+\frac{w_{s}D_{s}}{3R^2_{seed}}+w_{n}D_{n}}
\end{equation}
where $D_{c}, D_{s}, D_{n}$ are color/spatial/normal Euclidean distances. We assign the point with smallest distance into the superpoint associated with the current center. Iteratively, we set the newly added points as new centers to increase the superpoint until it meets the sphere boundary. In our experiments, the $w_{c}, w_{s}, w_{n}$ are set as 0.2, 0.4, 1. The interval of 50cm is the main parameter controlling the superpoint size, we add ablation studies in the main  paper Table \ref{tab:ablative}. Figure \ref{fig:app_vccs_regionGrow} shows an example of initial superpoints obtained by VCCS.

In implementation, we simply use the existing Point Cloud Library: 
\textit{\urlstyle{sf}\url{https://pcl.readthedocs.io/projects/tutorials/en/latest/supervoxel_clustering.html}}

\textbf{Details of Region Growing Algorithm}\cite{Adams1994}: This algorithm aims to merge points that are close enough in terms of local smoothness. The output of this algorithm is a set of clusters, where each cluster is a set of points that are considered to be a part of same smooth surface. The evaluation of smoothness is based on the similarity of point normals. Besides, the Region Growing algorithm begins its growth from the point that has the minimum curvature value. This is because the point with the minimum curvature is usually located in the flat area. Figure \ref{fig:app_vccs_regionGrow} shows an example. 

In implementation, we simply use the existing Point Cloud Library: \textit{\urlstyle{sf}\url{ https://pcl.readthedocs.io/projects/tutorials/en/latest/region_growing_segmentation.html}}

For an input point cloud, both VCCS and the Region Growing algorithm split it into a large number of partitions. We then combine these partitions together. Specifically, for each partition obtained by VCCS, if half of its 3D points are included by a partition obtained by Region Growing, we then merge all points in the former partition to the latter. 
\newline
\newline

\begin{table*}[t!] 
\centering
\caption{Per-category quantitative results on the validation split of ScanNet dataset.}
\label{tab:scannet_val_detail}
\resizebox{\textwidth}{!}
{
\begin{tabular}{crcccccccccccccccccccccccccccc}
\toprule[1.0pt]
& & OA(\%) & mAcc(\%)& mIoU(\%) & wall. & floor. & cab. & bed. & chair. & sofa. & table & door. & wind. & books. & pic. & counter. & desk. & curtain. & fridge. & shower. & toilet. & sink. & bathtub. & otherf.\\
\toprule[1.0pt]
\multirow{8}{*} 
& RandCNN 
&11.9$\pm$0.4 &8.4$\pm$0.1 &3.2$\pm$0 &9.3$\pm$0.4 &10.0$\pm$0.5 &3.5$\pm$0.5 &2.5$\pm$0.7 &6.8$\pm$0.9 &2.0$\pm$0.4 &4.8$\pm$0.7 &5.1$\pm$0.3 &3.9$\pm$0.2 &3.1$\pm$1.0 &1.7$\pm$0.5 &0.8$\pm$0.3 &2.3$\pm$0.4 &2.8$\pm$0.3 &1.0$\pm$0.1 &0.2$\pm$0.2 &0.2$\pm$0.1 &0.1$\pm$0.1 &0.6$\pm$0.1 &3.6$\pm$0.5\\

& van Kmeans 
&10.1$\pm$0.1 &10.0$\pm$0.1 &3.4$\pm$0 &9.0$\pm$0.4 &9.8$\pm$1.1 &3.2$\pm$0.6 &2.9$\pm$0.2 &5.5$\pm$0.4 &3.3$\pm$0.1 &4.3$\pm$0.2 &3.5$\pm$0.5 &5.5$\pm$0.4 &3.3$\pm$0.5 &\textbf{2.6}$\pm$0.4 &0.8$\pm$0.5 &2.9$\pm$0.4 &4.3$\pm$0.7 &0.8$\pm$0.2 &0.8$\pm$0.4 &0.7$\pm$0 &0.3$\pm$0.2 &0.9$\pm$0.2 &4.0$\pm$0.4\\

& van Kmeans-S 
&10.2$\pm$0.1 &9.8$\pm$0.3 &3.4$\pm$0.1 &8.9$\pm$0.5 &10.3$\pm$0.6 &3.4$\pm$0.3 &3.2$\pm$0.3 &5.5$\pm$0.2 &3.4$\pm$0.2 &4.2$\pm$0.6 &3.4$\pm$0.1 &5.2$\pm$1.4 &3.1$\pm$0.7 &2.6$\pm$0.4 &0.7$\pm$0.2 &2.8$\pm$0.4 &4.2$\pm$0.3 &0.6$\pm$0.2 &0.8$\pm$0.5 &0.7$\pm$0 &0.2$\pm$0.1 &1.0$\pm$0.2 &4.1$\pm$0.3\\

& van Kmeans-PFH
&10.4$\pm$0.2 &10.3$\pm$0.7 &3.5$\pm$0.2 &8.6$\pm$0.6 &12.7$\pm$0.1 &2.9$\pm$0.2 &2.8$\pm$0.1 &4.5$\pm$0.1 &3.2$\pm$0.1 &3.6$\pm$0.2 &3.7$\pm$0.3 &6.3$\pm$0.1 &4.0$\pm$0.3 &2.4$\pm$0.4 &1.0$\pm$0.1 &2.9$\pm$0.1 &3.2$\pm$0.8 &1.0$\pm$0.3 &1.0$\pm$0.4 &0.6$\pm$0.2 &0.4$\pm$0.1 &1.1$\pm$0.7 &3.5$\pm$0.5\\

& van Kmeans-S-PFH 
&12.2$\pm$0.6 &9.3$\pm$0.5 &3.6$\pm$0.1 &11.3$\pm$0.4 &12.3$\pm$1.4 &2.9$\pm$1.0 &2.4$\pm$0.6 &5.4$\pm$0.8 &2.8$\pm$0.4 &4.2$\pm$0.8 &3.8$\pm$0.5 &5.8$\pm$0.7 &3.8$\pm$0.6 &2.3$\pm$0.7 &1.2$\pm$0.3 &2.4$\pm$0.4 &2.9$\pm$1.7 &0.9$\pm$0.4 &1.4$\pm$0.7 &0.6$\pm$0.1 &0.1$\pm$0.1 &1.1$\pm$0.3 &4.1$\pm$0.7\\

& IIC \cite{Ji2019} 
&27.7$\pm$2.7 &6.1$\pm$1.2 &2.9$\pm$0.8 &25.3$\pm$3.9 &20.5$\pm$2.6 &0.6$\pm$1.0 &0.3$\pm$0.4 &3.7$\pm$4.7 &0.4$\pm$0.6 &1.3$\pm$1.6 &1.3$\pm$1.4 &1.1$\pm$1.5 &1.9$\pm$2.6 &0.2$\pm$0.1 &0.1$\pm$0.2 &0.6$\pm$0.8 &0.3$\pm$0.4 &0.4$\pm$0.6 &0$\pm$0 &0$\pm$0 &0$\pm$0 &0.2$\pm$0.3 &0.5$\pm$0.6\\

& IIC-S \cite{Ji2019} 
&18.3$\pm$2.6 &6.7$\pm$0.6 &3.4$\pm$0.1 &18.3$\pm$2.6 &16.0$\pm$1.6 &2.6$\pm$0.9 &2.3$\pm$0.4 &4.4$\pm$1.2 &2.0$\pm$0.3 &5.4$\pm$2.0 &3.2$\pm$1.6 &2.9$\pm$0.8 &3.3$\pm$1.4 &0.7$\pm$0.1 &0.4$\pm$0.2 &1.4$\pm$0.6 &1.6$\pm$0.7 &0.7$\pm$0.2 &0.1$\pm$0.2 &0.3$\pm$0.2 &0.1$\pm$0.1 &0$\pm$0 &2.6$\pm$0.8\\

& IIC-PFH \cite{Ji2019} 
&25.4$\pm$0.1 &6.3$\pm$0 &3.4$\pm$0 &29.6$\pm$0.2 &14.9$\pm$0.2 &1.1$\pm$0.1 &1.0$\pm$0 &5.6$\pm$0 &0.8$\pm$0.1 &3.6$\pm$0 &3.0$\pm$0 &1.6$\pm$0 &1.3$\pm$0.1 &0$\pm$0 &0.3$\pm$0.1 &1.0$\pm$0 &0.4$\pm$0 &0.4$\pm$0 &0.2$\pm$0 &0$\pm$0 &0.1$\pm$0 &0$\pm$0 &3.2$\pm$0.1\\

& IIC-S-PFH \cite{Ji2019} 
&18.9$\pm$0.3 &6.3$\pm$0.2 &3.0$\pm$0.1 &18.0$\pm$0.1 &15.9$\pm$0.3 &3.4$\pm$0.2 &0.9$\pm$0.1 &7.1$\pm$1.4 &0.6$\pm$0 &0.8$\pm$0.2 &4.3$\pm$0 &1.6$\pm$1.6 &3.5$\pm$0.1 &0.4$\pm$0 &0.1$\pm$0 &0.3$\pm$0.1 &0.3$\pm$0 &0$\pm$0 &0$\pm$0 &0.1$\pm$0 &0$\pm$0 &0$\pm$0 &2.7$\pm$1.6\\

& PICIE \cite{Cho2021} 
&20.4$\pm$0.5  &16.5$\pm$0.3 &7.6$\pm$0 &14.7$\pm$0.5 &24.5$\pm$1.8 &6.3$\pm$0.2 &5.2$\pm$1.8 &18.0$\pm$3.3 &8.4$\pm$1.3 &33.2$\pm$1.2 &6.7$\pm$0.5 &4.8$\pm$0.3 &9.3$\pm$4.0 &2.1$\pm$0.7 &0.1$\pm$0.1 &2.7$\pm$1.1 &8.0$\pm$1.3 &1.1$\pm$0.2 &2.1$\pm$1.8 &0$\pm$0 &0$\pm$0 &0.5$\pm$0.5 &5.0$\pm$0.3\\

& PICIE-S \cite{Cho2021} 
&35.6$\pm$1.1 &13.7$\pm$1.5 &8.1$\pm$0.5 &38.4$\pm$0.8 &53.9$\pm$0.8 &4.3$\pm$0.3 &2.7$\pm$0.3 &10.2$\pm$1.0 &6.3$\pm$0.8 &14.1$\pm$1.4 &5.2$\pm$0.6 &4.0$\pm$0  &6.0$\pm$0.2 &0.2$\pm$0.3 &1.3$\pm$0 &2.1$\pm$0.7 &1.5$\pm$0 &0.2$\pm$0.2 &0$\pm$0 &2.6$\pm$0.1 &3.1$\pm$0.7 &1.3$\pm$0 &4.3$\pm$0.2\\

& PICIE-PFH \cite{Cho2021} 
&33.1$\pm$1.4 &14.0$\pm$0.1 &8.1$\pm$0.3 &34.7$\pm$1.1 &54.8$\pm$3.0 &3.9$\pm$0.3 &5.4$\pm$1.9 &13.3$\pm$4.1 &6.5$\pm$1.6 &11.7$\pm$2.4 &4.2$\pm$0.4  &3.8$\pm$0.2 &6.5$\pm$1.1 &0.5$\pm$0.1 &1.0$\pm$0.6 &2.6$\pm$1.5 &5.0$\pm$0.4 &\textbf{1.3}$\pm$0.3 &1.0$\pm$0.9 &0.6$\pm$0.8 &0$\pm$0 &0.7$\pm$0.1 &4.4$\pm$0.6\\

& PICIE-S-PFH \cite{Cho2021} 
&23.6$\pm$0.4 &15.1$\pm$0.6 &7.4$\pm$0.2 &18.1$\pm$0.8 &39.1$\pm$1.5 &5.4$\pm$0.2 &4.9$\pm$0.4 &13.4$\pm$0.9 &6.9$\pm$0.4 &20.3$\pm$5.8 &5.8$\pm$0.1 &4.5$\pm$0.3  &7.7$\pm$0.5 &1.2$\pm$1.0 &3.0$\pm$1.9 &5.8$\pm$0.7 &4.7$\pm$0.8 &0.6$\pm$0.5 &1.2$\pm$1.0 &0.4$\pm$0.3 &0$\pm$0 &1.1$\pm$0.3 &4.5$\pm$0.2\\

& \textbf{\nickname{}(Ours)} &\textbf{57.3}$\pm$2.3 &\textbf{44.2}$\pm$3.1 &\textbf{25.4}$\pm$2.3 &\textbf{40.7}$\pm$2.0 &\textbf{89.8}$\pm$0.4 &\textbf{24.0}$\pm$5.8 &\textbf{47.2}$\pm$2.0 &\textbf{45.5}$\pm$19.0 &\textbf{43.0}$\pm$1.4 &\textbf{39.4}$\pm$3.4 &\textbf{14.1}$\pm$0.5 &\textbf{20.0}$\pm$0.3 &\textbf{53.5}$\pm$6.6 &0.1$\pm$0.1 &\textbf{5.4}$\pm$9.5 &\textbf{13.3}$\pm$0.5 &\textbf{8.4}$\pm$0.8 &\textbf{2.1}$\pm$0.6 &\textbf{11.3}$\pm$1.2 &\textbf{20.6}$\pm$18.2 &\textbf{19.4}$\pm$1.2 &0$\pm$0 &\textbf{9.8}$\pm$2.7\\
\bottomrule[1.0pt]
\end{tabular}
}
\end{table*}

\begin{table*}[h!] 
\centering
\caption{Per-category quantitative results on the hidden test split of ScanNet dataset.}
\label{tab:scannet_test_detail}
\resizebox{\textwidth}{!}
{
\begin{tabular}{crcccccccccccccccccccccccccccc}
\toprule[1.0pt]
& &mIoU(\%) & wall. & floor. & cab. & bed. & chair. & sofa. & table & door. & wind. & books. & pic. & counter. & desk. & curtain. & fridge. & shower. & toilet. & sink. & bathtub. & otherf.\\
\toprule[1.0pt]
\multirow{3}{*}{\makecell[c]{Supervised\\Methods} }
& PointNet++ \cite{Qi2017} &33.9 &52.3 &67.7 &25.6 &47.8 &36 &34.6 &23.2 &26.1 &25.2 &45.8 &11.7 &25.0 &27.8 &24.7 &18.3 &14.5 &54.8 &36.4 &58.4 &18.3\\
& DGCNN \cite{Wang2018c} &44.6 &72.3 &93.7 &36.6 &62.3 &65.1 &57.7 &44.5 &33.0 &39.4 &46.3 &12.6 &31.0 &34.9 &38.9 &28.5 &22.4 &62.5 &35.0 &47.4 &27.1\\
& PointCNN \cite{Li2018f}&45.8 &70.9 &94.4 &32.1 &61.1 &71.5 &54.5 &45.6 &31.9 &47.5 &35.6 &16.4 &29.9 &32.8 &37.6 &21.6 &22.9 &75.5 &48.4 &57.7 &28.5\\
& SparseConv \cite{Graham2018} &72.5 &86.5 &95.5 &72.1 &82.1 &86.9 &82.3 &62.8 &61.4 &68.3 &84.6 &32.5 &53.3 &60.3 &75.4 &71.0 &87.0 &93.4 &72.4 &64.7 &57.2\\
\toprule[1.0pt]
\multirow{1}{*}{\makecell[c]{Unsupervised Methods} } 
& \textbf{\nickname{}(Ours)} &26.9 &32.8 &89.6 &15.2 &62.9 &55.3 &38.9 &32.0 &14.4 &23.0 &59.9 &0 &12.5 &11.4 &6.1 &1.2 &9.3 &43.9 &14.0 &0 &16.5\\
\bottomrule[1.0pt]
\end{tabular}
}
\end{table*}

\vspace{-1.3cm}
\section{Point Feature Histograms Descriptors}
\label{sec:app_PFH}

In our semantic primitive clustering module, we adopt PFH feature \cite{Rusu2008} which explicitly measures the surface normal distributions to augment neural features for better semantic primitive clustering. 

Given a center point with its neighbouring points, PFH iteratively selects any two of them and computes a set of angles from point normals. In our implementation, we adopt a simplified version to save computation. Specifically, for a superpoint that contains a set of points, we compute the cosine distances between normals of any two points in this suerpoint, and compute the distribution of cosine distances to form a histogram in the range $\left[-1, 1\right]$ with an interval 0.2. In this way, we can get a 10-dimensional vector to describe the normal distribution of a superpoint, and regard it an additional features for semantic primitives clustering.

\begin{figure*}[thb]
\centering
   \includegraphics[width=1\linewidth]{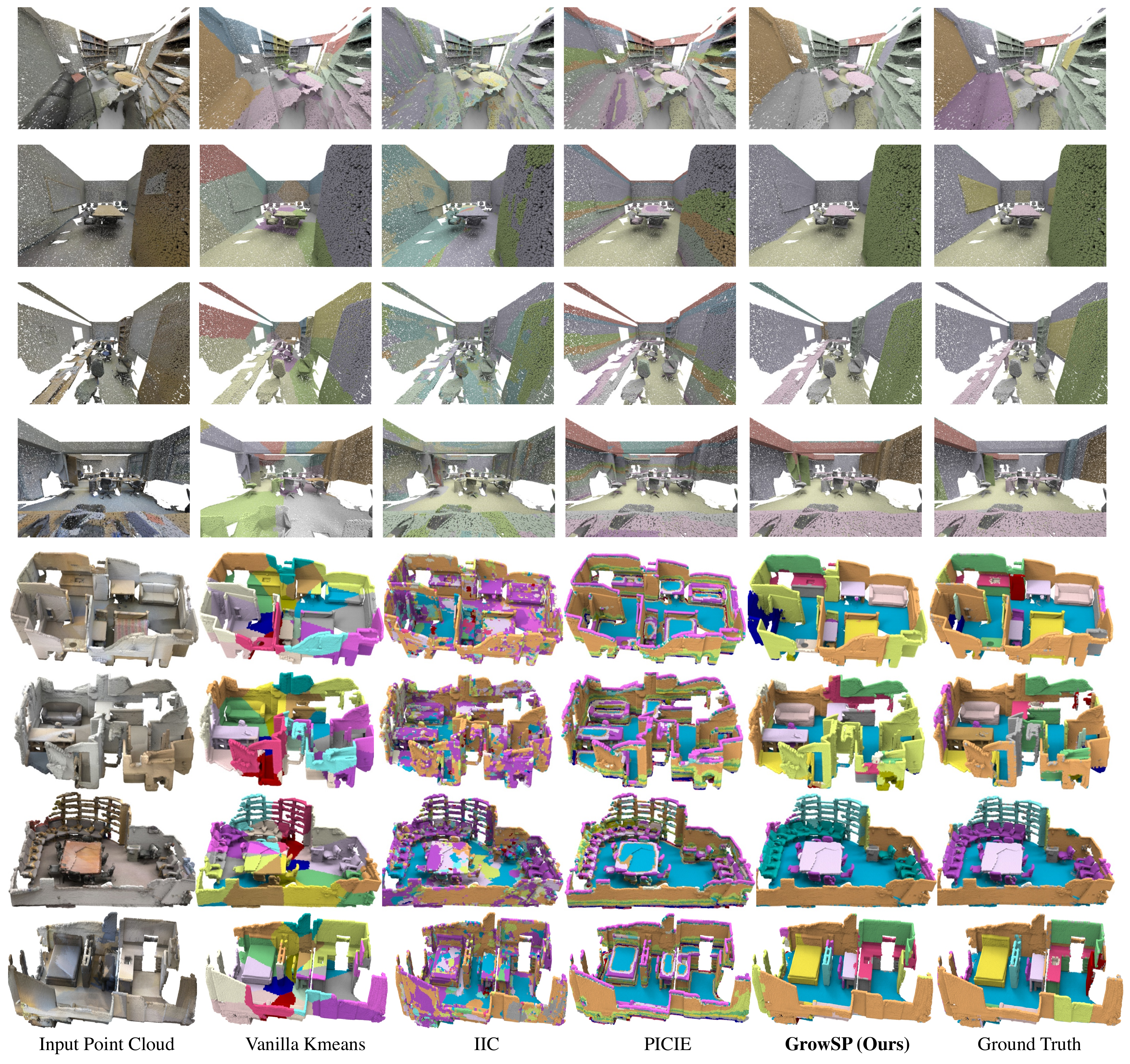}
\caption{The top four rows show qualitative results for S3DIS, the bottom four rows for ScanNet.}
\label{fig:app_s3dis_scannet}
\end{figure*}

\section{Evaluation on S3DIS}
\label{sec:app_s3dis}

In S3DIS, we use the same hyper-parameters for all 6 Areas. Following SparseConv \cite{Graham2018}, we firstly use a grid size of 0.01m to sub-grid downsample input point clouds. We then use a 5cm voxel size to voxelize each point cloud. 
The cross-entropy loss is minimized by SGD optimizer with a batch size of 10, a momentum of 0.9, an initial learning rate of 0.1. The learning rate is decreased by  Poly scheduler, and we train 1300 epochs for each area of S3DIS. During training, we cluster 300 semantic primitives for all epochs. The semantic primitive clustering, superpoint growing, and pseudo labels updating are conducted every 10 epochs.

Because the category \textit{clutter} does not have consistent geometry, we exclude it during training. We never use these points to compute losses or apply K-means. \cref{tab:s3dis_a1,tab:s3dis_a2,tab:s3dis_a3,tab:s3dis_a4,tab:s3dis_a5,tab:s3dis_a6} present the per-category results on Area-1/2/3/4/5/6. Our \textbf{\nickname} consistently outperforms all baselines and achieve comparable performance with the fully-superviesed PointNet \cite{Qi2016}. More qualitative results can be seen in Figure \ref{fig:app_s3dis_scannet}.

\section{Evaluation on ScanNet}
\label{sec:app_scannet}

In ScanNet, the hyper-parameters are the same as in S3DIS. We also use a 5cm voxel size to voxelize each point cloud.
The cross-entropy loss is minimized by SGD optimizer with a batch size of 8, a momentum of 0.9, and an initial learning rate of 0.1. The learning rate is decreased by Poly scheduler. We train a total of 800 epochs. 
Since ScanNet has an \textit{undefined} category which does not have consistent geometry, we exclude(masked) these points during training. We feed these points into the network but neither compute their losses nor apply K-means to them.

The per-category results on both validation and hidden test sets  are presented in Tables \ref{tab:scannet_val_detail} \& \ref{tab:scannet_test_detail}. Our method outperforms all baselines and achieve comparable results with PointNet++ \cite{Qi2017}. Figure \ref{fig:app_s3dis_scannet} shows more qualitative results.

\section{Generalization to Unseen Datasets}

In this section, we further evaluate whether the learned features of our unsupervised method are indeed general to unseen scenes. In particular, having the well-trained models of both S3DIS and ScanNet in Sections \ref{sec:exp_s3dis}\&\ref{sec:exp_scannet}, we conduct the following two groups of experiments.
\begin{itemize}[leftmargin=*]
\setlength{\itemsep}{1pt}
\setlength{\parsep}{1pt}
\setlength{\parskip}{1pt}
    \item \textit{Group 1: Generalization from ScanNet to S3DIS}. We directly use the well-trained model of ScanNet to test on 6 areas of S3DIS. The final classifier uses the 20 centroids estimated by K-means on the training split of ScanNet. This means that, the neural network is completely unaware of any information of S3DIS dataset.   
    \item \textit{Group 2: Generalization from S3DIS to ScanNet}. We directly use the well-trained 6 models of S3DIS to test on the validation split of ScanNet. 
\end{itemize}

\textbf{Analysis:} Table \ref{tab:app_scannet_2_s3dis} shows results of Group 1. We can see that our method trained on ScanNet dataset has achieved superior generalization capability to the unseen S3DIS dataset, with mIoU scores about 30\% in several areas, which are clearly better than all baselines. Notably, these scores are slightly higher than the results in Tables \ref{tab:exp_scannet_val}\&\ref{tab:exp_scannet_online}. We speculate it is because ScanNet dataset itself consists of very diverse geometries and semantic elements which are well discovered by our model, whereas S3DIS dataset consists of fewer semantic elements and is relatively easier. Table \ref{tab:app_s3dis_2_scannet} shows results of Group 2. It can be seen that the generalization performance of our method from S3DIS to ScanNet is clearly better than baselines, although the overall capability is weaker than that of Group 1 because of the relatively simple rooms in S3DIS dataset for training. 

\begin{table}[thb]
\centering
\caption{Generalization ability of models trained on ScanNet \cite{Dai2017} to the unseen 6 areas of S3DIS \cite{Armeni2017}. The mIoU scores with standard deviations of 12 categories are reported.}
\label{tab:app_scannet_2_s3dis}
\resizebox{0.44\textwidth}{!}{
\begin{tabular}{rcccccc}
\toprule[1.0pt]
 \textit{test on $\rightarrow$} & Area-1 & Area-2  & Area-3 & Area-4 & Area-5  & Area-6  \\
\toprule[1.0pt]
    IIC \cite{Ji2019} &3.7$\pm$0.5 &3.8$\pm$0.4 &3.8$\pm$0.2 &4.0$\pm$0.5 &3.8$\pm$0.2 &3.7$\pm$0.4 \\
    IIC-S \cite{Ji2019} &6.7$\pm$0.1 &5.7$\pm$0 &6.4$\pm$0.2 &5.8$\pm$0 &5.9$\pm$0 &6.5$\pm$0.1 \\
    PICIE \cite{Cho2021} &13.5$\pm$0.1 &12.7$\pm$0.2 &13.4$\pm$0.1 &12.8$\pm$0.1 &11.3$\pm$0.4 &13.1$\pm$0.1  \\
    PICIE-S \cite{Cho2021} &14.7$\pm$0.9 &13.9$\pm$0.8 &15.1$\pm$0.7 &14.7$\pm$0.4 &14.2$\pm$0.3 &15.8$\pm$0.2  \\
\textbf{\nickname{} (Ours)} & \textbf{24.2}$\pm$1.9 &\textbf{21.9}$\pm$1.7 &\textbf{26.1}$\pm$2.8 &\textbf{25.0}$\pm$2.8 &\textbf{23.7}$\pm$2.3 &\textbf{27.9}$\pm$2.5  \\
\toprule[1.0pt]
\end{tabular}
}\vspace{-0.2cm}
\end{table}

\begin{table}[thb]
\centering
\caption{Generalization ability of models trained on different areas of S3DIS \cite{Armeni2017} to the unseen val split of ScanNet \cite{Dai2017}. The mIoU scores with standard deviations of 20 categories are reported.}
\label{tab:app_s3dis_2_scannet}
\resizebox{0.44\textwidth}{!}{
\begin{tabular}{rcccccc}
\toprule[1.0pt]
 \multirow{2}{*}{\makecell[c]{\textit{model trained on $\rightarrow$}}} & \multirow{2}{*}{\makecell[c]{Areas\\2/3/4/5/6}} & \multirow{2}{*}{\makecell[c]{Areas\\1/3/4/5/6}} & \multirow{2}{*}{\makecell[c]{Areas\\1/2/4/5/6}} & \multirow{2}{*}{\makecell[c]{Areas\\1/2/3/5/6}} & \multirow{2}{*}{\makecell[c]{Areas\\1/2/3/4/6}}  & \multirow{2}{*}{\makecell[c]{Areas\\1/2/3/4/5}}  \\
  &   &   &   &   &   &   \\
\toprule[1.0pt]
    IIC \cite{Ji2019} &3.5$\pm$0 &3.4$\pm$0 &3.7$\pm$0.1 &3.5$\pm$0.1 &3.5$\pm$0 &3.6$\pm$0 \\
    IIC-S \cite{Ji2019} &3.9$\pm$0.1 &3.9$\pm$0.1 &4.0$\pm$0.1 &3.9$\pm$0 &3.9$\pm$0.1 &3.9$\pm$0 \\
    PICIE \cite{Cho2021} &5.6$\pm$0.2 &5.1$\pm$0.1 &5.0$\pm$0.1 &5.9$\pm$0.3 &6.0$\pm$0.3 &5.5$\pm$0.2  \\
    PICIE-S \cite{Cho2021} &6.9$\pm$0.3 &6.9$\pm$0.7 &6.9$\pm$0.8 &8.1$\pm$0.4 &8.4$\pm$0.3 &6.7$\pm$0.9  \\
\textbf{\nickname{} (Ours)} &\textbf{16.9}$\pm$0.6 & \textbf{17.8}$\pm$0.6 &\textbf{16.4}$\pm$0.5 &\textbf{16.1}$\pm$0.6 &\textbf{17.1}$\pm$0.8 &\textbf{15.3}$\pm$0.3  \\
\toprule[1.0pt]
\end{tabular}
}\vspace{-0.2cm}
\end{table}

\section{Evaluation on SemanticKITTI}
\label{sec:app_semantickitti}

Considering that outdoor point clouds are usually dominated by \textit{road} and the point density is significantly different from that of indoor datasets, we opt for (RANSAC $+$ Euclidean Clustering) as an alternative to (VCCS $+$ region growing) to construct initial superpoints, 

\textbf{RANSAC:} In our experiment, we choose to fit a plane by RANSAC and take points with a distance smaller than 0.2m as a huge superpoint.

\textbf{Euclidean Clustering:} After fitting the largest plane which normally corresponds to \textit{road}, we construct initial superpoints for remaining points by Euclidean clustering. Specifically, if the Euclidean distance of two points is smaller than 0.2m, they are assigned into the same superpoint; otherwise not.

\textbf{Training/Testing Details:} In training, we voxelize point clouds by a grid size of 15cm without any other pre-processing. Following Mix3D \cite{Nekrasov2021}, we use an AdamW optimizer with a batchsize of 16 to train 400 epochs. The learning rate is decreased by OneCycleLR with a max learning rate of 0.01. The number of primitive $S$ is set as 500, $M^1$ as 80, and $M^T$ as 30. The semantic primitive clustering, superpoint growing, and pseudo labels updating are conducted every 10 epochs. Since SemanticKITTI has nearly 20000 training scans and each covers an extremely spacious range, we only randomly select 1500 scenes in each round (10 epochs) and crop each point cloud by a 50m radius sphere to train the network for efficiency. Similarly, the \textit{undefined background} points do not have consistent geometry, so we exclude(masked) these points during training. We feed these points into the network but neither compute their losses nor apply K-means on them. During testing, all raw points are fed into the network for evaluation.

\begin{figure*}[h!]
\centering
   \includegraphics[width=1.0\linewidth]{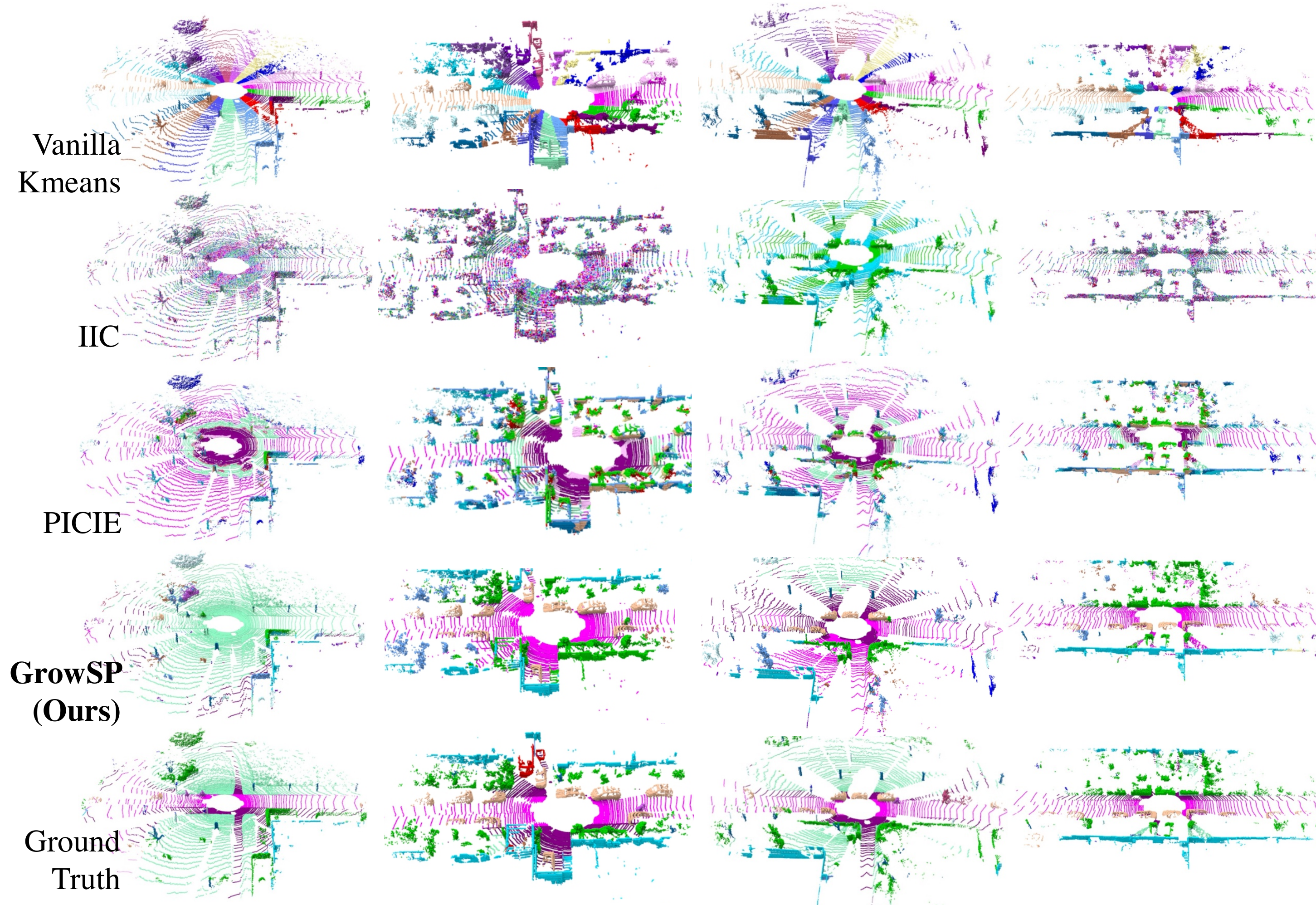}
\caption{Qualitative results on SemanticKITTI dataset.}
\label{fig:vis_kitti}
\end{figure*}

Tables \ref{tab:kitti_val}\&\ref{tab:kitti_test} show the per-category results on both validation and hidden test sets, and our methods achieve SOTA compared with other unsupervised approaches. Figure \ref{fig:vis_kitti} shows qualitative results. \\

More qualitative results and video demo can be found at \url{https://github.com/vLAR-group/GrowSP}

\begin{table*}[t!] 
\centering
\caption{Per-category quantitative results on the \textbf{offline validation split} of SemanticKITTI dataset.}
\label{tab:kitti_val}
\resizebox{\textwidth}{!}
{
\begin{tabular}{crcccccccccccccccccccccccccccc}
\toprule[1.0pt]
& & OA(\%) & mAcc(\%)& mIoU(\%) & car. & bike. & mbike. & truck. & vehicle. & person. & cyclist. & mcyclist. & road. & parking. & sidewalk. & other-gr. & building. & fence. & veget. & trunk. & terrain. & pole. & sign.\\
\toprule[1.0pt]
\multirow{8}{*} 
& RandCNN 
&25.4$\pm$3.3 &6.0$\pm$0.2 &3.3$\pm$0.1 &2.5$\pm$0.4 &0$\pm$0 &0$\pm$0 &0$\pm$0 &0.2$\pm$0.1 &0$\pm$0 &0$\pm$0 &0$\pm$0 &8.5$\pm$2.1 &0.8$\pm$0.5 &4.9$\pm$1.8 &0.3$\pm$0.3 &6.2$\pm$1.3 &1.3$\pm$0.3 &29.0$\pm$3.1 &1.0$\pm$0.2 &8.1$\pm$1.6 &0.4$\pm$0.1 &0.1$\pm$0\\

& van Kmeans 
&8.1$\pm$0 &8.2$\pm$0.1 &2.4$\pm$0 &5.6$\pm$0.2 &0.1$\pm$0 &0.1$\pm$0 &0.2$\pm$0 &0.5$\pm$0.1 &0.1$\pm$0 &0$\pm$0 &0$\pm$0 &12.3$\pm$0.1 &1.1$\pm$0.1 &4.4$\pm$0.1 &0.3$\pm$0 &5.8$\pm$0.2 &2.0$\pm$0 &5.7$\pm$0.1 &1.4$\pm$0 &5.0$\pm$0.1 &0.5$\pm$0 &0.1$\pm$0\\

& van Kmeans-S 
&10.3$\pm$0.3 &7.7$\pm$0.1 &2.6$\pm$0 &5.6$\pm$0.4 &0.1$\pm$0.1 &0.1$\pm$0.1 &0.1$\pm$0.1 &0.3$\pm$0 &0.1$\pm$0 &0$\pm$0 &0$\pm$0 &13.5$\pm$0.6 &1.0$\pm$0.4 &5.0$\pm$0.2 &0.3$\pm$0 &7.1$\pm$0.6 &1.5$\pm$0.2 &7.5$\pm$0.7 &1.5$\pm$0.1 &6.0$\pm$0.1 &0.4$\pm$0.1 &0.1$\pm$0\\

& van Kmeans-PFH
&11.2$\pm$0.6 &7.5$\pm$0.7 &2.7$\pm$0.1 &4.5$\pm$0.8 &0.1$\pm$0 &0.1$\pm$0.1 &0.2$\pm$0.1 &0.1$\pm$0.1 &0.2$\pm$0 &0.2$\pm$0.2 &0$\pm$0 &9.1$\pm$0.8 &1.6$\pm$0.6 &4.9$\pm$0.4 &0.2$\pm$0.1 &8.2$\pm$0.8 &1.6$\pm$0.2 &9.6$\pm$0.5 &1.4$\pm$0 &7.5$\pm$0.7 &0.3$\pm$0 &0.3$\pm$0.2\\

& van Kmeans-S-PFH
&13.2$\pm$1.8 &8.1$\pm$0.4 &3.0$\pm$0.2 &4.8$\pm$0.5 &0.1$\pm$0.1 &0.1$\pm$0.1 &0.3$\pm$0.2 &0.5$\pm$0.2 &0.1$\pm$0 &0.2$\pm$0.2 &0$\pm$0 &11.3$\pm$2.5 &1.5$\pm$0.5 &5.3$\pm$0.8 &0.3$\pm$0 &8.4$\pm$0.7 &1.5$\pm$0.2 &11.3$\pm$3.2 &1.5$\pm$0.1 &8.4$\pm$0.5 &0.4$\pm$0.1 &0.3$\pm$0.1\\

& IIC \cite{Ji2019} 
&26.2$\pm$1.5 &5.8$\pm$0.4 &3.1$\pm$0.3 &1.6$\pm$0.9 &0$\pm$0 &0$\pm$0 &0$\pm$0 &0$\pm$0 &0$\pm$0 &0$\pm$0  &0$\pm$0 &8.9$\pm$2.0 &0.1$\pm$0.1 &2.6$\pm$1.8 &0$\pm$0 &7.1$\pm$4.2 &0.2$\pm$0.1 &26.5$\pm$2.5 &0.3$\pm$0.4 &11.5$\pm$1.5 &0.1$\pm$0.1 &0.1$\pm$0.1\\

& IIC-S \cite{Ji2019} 
&23.9$\pm$1.1 &6.1$\pm$0.3 &3.2$\pm$0.2 &1.6$\pm$0.8 &0$\pm$0 &0$\pm$0 &0.1$\pm$0.1 &0.1$\pm$0.1 &0$\pm$0 &0.1$\pm$0.1 &\textbf{9.7}$\pm$1.9 &0.6$\pm$0.5 &\textbf{4.3}$\pm$2.8 &0.1$\pm$0.1 &\textbf{8.8}$\pm$3.2 &0.5$\pm$0.6 &\textbf{24.3}$\pm$2.3 &0.6$\pm$0.5 &9.7$\pm$2.6 &0.3$\pm$0.3 &0.1$\pm$0.1 &0$\pm$0.1\\

& IIC-PFH \cite{Ji2019} 
&20.1$\pm$0.1 &7.2$\pm$0.1 &3.6$\pm$0 &5.8$\pm$0 &0.1$\pm$0.1 &0.2$\pm$0 &0.2$\pm$0 &0.5$\pm$0 &0.2$\pm$0 &0.1$\pm$0 &0$\pm$0 &14.5$\pm$0.2 &1.1$\pm$0.3 &6.6$\pm$0.2 &0.1$\pm$0 &6.8$\pm$0.1 &1.6$\pm$0.2 &19.7$\pm$0 &2.1$\pm$0 &8.4$\pm$0.1 &0.6$\pm$0.1 &0.1$\pm$0\\

& IIC-S-PFH \cite{Ji2019} 
&23.4$\pm$0 &9.0$\pm$0 &4.6$\pm$0 &10.0$\pm$0.1 &0.1$\pm$0 &0$\pm$0 &0.3$\pm$0 &0.4$\pm$0 &\textbf{0.3}$\pm$0 &\textbf{0.3}$\pm$0 &0$\pm$0 &21.7$\pm$0.2 &2.4$\pm$0 &10.0$\pm$0.1 &0$\pm$0 &8.7$\pm$0 &1.6$\pm$0 &19.7$\pm$0.2 &1.1$\pm$0 &9.7$\pm$0.1 &0.4$\pm$0 &0.2$\pm$0\\

& PICIE \cite{Cho2021} 
&22.3$\pm$0.4  &14.6$\pm$0.3 &5.9$\pm$0.1 &\textbf{7.4}$\pm$0.2 &0.3$\pm$0.2 &0$\pm$0 &0.1$\pm$0 &0.6$\pm$0.1 &0.3$\pm$0.1 &0.1$\pm$0.1 &0$\pm$0 &26.5$\pm$0.3 &1.6$\pm$0.1 &14.8$\pm$1.4 &0.6$\pm$0.3 &20.5$\pm$0.4 &4.8$\pm$0.1 &16.3$\pm$1.0 &2.1$\pm$0.9 &14.2$\pm$0.9 &1.4$\pm$0.3 &0.4$\pm$0.2\\

& PICIE-S \cite{Cho2021} 
&18.4$\pm$0.5 &13.2$\pm$0.2 &5.1$\pm$0.1 &6.1$\pm$1.4 &0.1$\pm$0 &0$\pm$0 &0.1$\pm$0.1 &0.4$\pm$0.1 &0.3$\pm$0.1 &0.1$\pm$0.1 &0$\pm$0 &21.3$\pm$1.4  &1.7$\pm$0.1 &12.9$\pm$2.3 &0.4$\pm$0.2 &21.2$\pm$0.9 &2.6$\pm$0.3 &13.4$\pm$0.4 &2.4$\pm$0.3 &11.5$\pm$2.9 &\textbf{2.6}$\pm$0.2 &0.4$\pm$0\\

& PICIE-PFH \cite{Cho2021} 
&46.6$\pm$0.2 &10.1$\pm$0 &4.7$\pm$0 &0$\pm$0 &0$\pm$0 &0$\pm$0 &0$\pm$0 &0$\pm$0 &0$\pm$0 &0$\pm$0 &0$\pm$0 &39.7$\pm$0.8  &0$\pm$0 &0$\pm$0 &0$\pm$0 &0$\pm$0 &0$\pm$0 &50.2$\pm$0.3 &0$\pm$0 &0$\pm$0 &0$\pm$0 &0$\pm$0\\

& PICIE-S-PFH \cite{Cho2021} 
&42.7$\pm$2.1 &11.5$\pm$0.2 &6.8$\pm$0.6 &4.8$\pm$2.6 &0$\pm$0 &0$\pm$0 &0$\pm$0 &0$\pm$0 &0$\pm$0 &0$\pm$0 &0$\pm$0 &32.0$\pm$6.3  &0.6$\pm$1.0 &12.5$\pm$8.7 &0$\pm$0 &25.5$\pm$0.6 &0.8$\pm$1.0 &43.6$\pm$1.2 &0.5$\pm$0.4 &9.2$\pm$9.3 &0$\pm$0 &0$\pm$0\\

& \textbf{\nickname{}(Ours)} &\textbf{38.3}$\pm$1.0 &\textbf{19.7}$\pm$0.6 &\textbf{13.2}$\pm$0.1 &\textbf{76.0}$\pm$0.4 &0$\pm$0 &\textbf{0.4}$\pm$0.2 &\textbf{0.9}$\pm$0.7 &\textbf{1.0}$\pm$0.1 &0.1$\pm$0.2 &0.1$\pm$0.2 &0$\pm$0 &\textbf{26.8}$\pm$3.5 &1.0$\pm$0.4 &\textbf{13.8}$\pm$4.5 &0.4$\pm$0.3 &\textbf{39.2}$\pm$2.1 &1.3$\pm$0.4 &\textbf{26.7}$\pm$1.5 &\textbf{25.1}$\pm$0.7 &\textbf{35.5}$\pm$1.9 &0.2$\pm$0.1 &\textbf{2.1}$\pm$0.1\\
\bottomrule[1.0pt]
\end{tabular}
}
\end{table*}
\begin{table*}[h!] 
\centering
\caption{Per-category quantitative results on the \textbf{hidden test split} of SemanticKITTI dataset.}
\label{tab:kitti_test}
\resizebox{\textwidth}{!}
{
\begin{tabular}{crcccccccccccccccccccccccccccc}
\toprule[1.0pt]
&& mIoU(\%) & car. & bike. & mbike. & truck. & vehicle. & person. & cyclist. & mcyclist. & road. & parking. & sidewalk. & other-gr. & building. & fence. & veget. & trunk. & terrain. & pole. & sign.\\
\toprule[1.0pt]
\multirow{3}{*}{\makecell[c]{Supervised\\Methods} }
& PointNet \cite{Qi2016} &14.6 &46.3 &1.3 &0.3 &4.6 &0.8 &0.2 &0.2 &0 &61.6 &15.8 &35.7 &1.4 &41.4 &12.9 &31.0 &4.6 &17.6 &2.4 &3.7\\
& PointNet++ \cite{Qi2017} &20.1 &53.7 &0.9 &0.2 &0.9 &0.2 &0.9 &1.0 &0 &72.0 &18.7 &41.8 &5.6 &62.3 &16.9 &46.5 &13.8 &30.0 &6.0 &8.9\\
& SparseConv \cite{Choy2019} &53.2 &94.0 &26.4 &24.5 &27.5 &18.4 &40.5 &46.7 &13.5 &88.4 &57.1 &71.4 &22.6 &90.4 &62.5 &83.5 &65.3 &65.8 &54.0 &59.1\\
\toprule[1.0pt]
\multirow{1}{*}{\makecell[c]{Unsupervised Methods} } 
& \textbf{\nickname{}(Ours)} &14.3 &81.9 &0.1 &0.5 &0.2 &1.0 &0.3 &0 &0 &25.0 &0 &17.4 &0.5 &64.6 &1.4 &29.4 &26.6 &22.4 &0.3 &0.5\\
\bottomrule[1.0pt]
\end{tabular}
}
\end{table*}

\end{document}